\documentclass[preprint,3p,times]{elsarticle}

\usepackage{amssymb,amsthm,amsmath,longtable,array}
\usepackage{tikz,circuitikz,graphicx,subcaption}
\usepackage[linesnumbered,lined,ruled]{algorithm2e}

\usepackage{tabularx,booktabs,ragged2e,float}
\newcolumntype{L}{>{\RaggedRight\arraybackslash}X} % Define a left-aligned X column

\usepackage[normalem]{ulem}
\usepackage{xcolor}
\usepackage{lineno}
\DeclareMathOperator*{\argmax}{arg\,max}

\journal{Journal of Intelligent Manufacturing}

\begin{document}

\begin{frontmatter}

\title{Enhancing Obsolescence Forecasting with Deep Generative Data Augmentation: A Semi-Supervised Framework for Low-Data Industrial Applications\tnoteref{titlenote}}

\tnotetext[titlenote]{This work was supported by the Société Nationale des Chemins de fer Français (SNCF) Réseau (French National Railway Company).}

\author[quartz,sncf]{Elie Saad}
\ead{elie.saad@isae-supmeca.fr, elie.saad@Réseau.sncf.fr}
\ead[url]{https://eliesaad.net/}
\author[quartz]{Mariem Besbes}
\author[quartz]{Marc Zolghadri}
\author[sncf]{Victor Czmil}
\author[laas]{Claude Baron}
\author[sncf]{Vincent Bourgeois}

\affiliation[quartz]{organization={Quartz Laboratory, ISAE-Supméca},
            addressline={3 Rue Fernand Hainaut}, 
            city={Saint-Ouen-sur-Seine},
            postcode={93400}, 
            country={France}}
\affiliation[sncf]{organization={SNCF Réseau},
            addressline={6 Av. François Mitterrand}, 
            city={Saint-Denis},
            postcode={93210}, 
            country={France}}
\affiliation[laas]{organization={LAAS-CNRS},
            addressline={7 Av. du Colonel Roche}, 
            city={Toulouse},
            postcode={31400}, 
            country={France}}

\begin{abstract}
The challenge of electronic component obsolescence is particularly critical  in systems with long life cycles. Various obsolescence management methods are employed to mitigate its impact, with obsolescence forecasting being a highly sought-after and prominent approach. As a result, numerous machine learning-based forecasting methods have been proposed. However, machine learning models require a substantial amount of relevant data to achieve high precision, which is lacking in the current obsolescence landscape in some situations. This work introduces a novel framework for obsolescence forecasting based on deep learning. The proposed framework solves the lack of available data through deep generative modeling, where new obsolescence cases are generated and used to augment the training dataset. The augmented dataset is then used to train a classical machine learning-based obsolescence forecasting model. To train classical forecasting models using augmented datasets, existing classical supervised-learning classifiers are adapted for semi-supervised learning within this framework. The proposed framework demonstrates state-of-the-art results on benchmarking datasets. 
\end{abstract}


\end{frontmatter}

%% main text
\section{Introduction}\label{sec:introduction}
Systems are vulnerable to obsolescence, which can affect parts, modules, components, equipment, or even entire systems as identified by the Department of Defense (DoD) \citep{SD22}. Obsolescence is an unavoidable reality \citep{trabelsi2021prediction}, influenced by factors such as rapidly changing consumer preferences, technological innovation, and wear and tear among others. These elements interact to influence the perceived and actual lifespan of products and technologies, as described by \citet{rojo2012}. Obsolescence refers to being outdated, useless, or unused. According to \citet{IEC62402}, obsolescence is the transition from the state of availability to the state of unavailability of an entity from its manufacturer in accordance with the original specification. In American literature, and in particular throughout DoD documentation, the term “Diminishing Manufacturing Sources and Material Shortages” (DMSMS) is commonly used to refer to obsolescence as noted by \citet{zolghadri2023micro}. A semantic distinction can be made between obsolescence, which refers to the changing needs of a product in its current state, and scarcity, which concerns the observed or imminent interruption of the supply of goods (sources) and resources needed for its production. As \citet{mellal2020obsolescence} points out, there is currently a lack of consensus and clarity in the literature regarding the definition and understanding of obsolescence. Therefore, the definition of obsolescence as described by \citet{zolghadri2021obsolescence} will be used within this paper, which states that the term obsolescence is the process of becoming or achieving a state of unavailability, inadequacy, unsuitability, or any of the previous states combined. 

Obsolescence can occur when a product loses value over time due to wear and tear, technological change or trends in consumer behaviour \citep{zrobek2011remarks,Butt2015ObsolescenceTA,rust2022literature}. Other types of obsolescence include (1) economic obsolescence, which occurs when a product is no longer economically feasible or desirable due to shifts in the market or the economy \citep{Butt2015ObsolescenceTA}, (2) design obsolescence, which occurs when a product becomes outdated due to design or engineering modifications \citep{rust2022literature}, (3) system obsolescence, which occurs when a product becomes incongruent with other products or systems \citep{rust2022literature}, and (4) component obsolescence, which occurs when a product contains at least one component that becomes obsolete or unsupported, putting the entire system at risk of unavailability. \citep{rust2022literature}.  

Obsolescence has a negative economic impact as it increases costs due to the need to design and implement mitigation solutions, thus reducing the net margin of companies \citep{Butt2015ObsolescenceTA,rust2022literature,rojo2012}. However, it can lead to new businesses servicing or maintaining legacy systems that are becoming obsolete.

From a technological perspective, obsolescence can also have a negative impact, such as various technologies becoming incompatible, which can lead to higher expenses and reduced effectiveness \citep{Butt2015ObsolescenceTA,rust2022literature,rojo2012}. However, obsolescence can be an innovation and technological advancement driver which fosters economic expansion, as well as enhanced efficiency and productivity, as outdated technologies are substituted with newer, more advanced alternatives \citep{rust2022literature}. Systems often represent a significant and long-term investment with a planned service life spanning several decades, while most of their components and/or their technologies have considerably shorter lifespans. This constitutes one of the primary causes of obsolescence cases \citep{Jenab2014ObsolescenceMI,DeFrancesco2019UseOT,riascos2019aging}. This disparity creates substantial challenges for system managers, who must anticipate and manage these risks to ensure system availability throughout their operational life.

Managing obsolescence efficiently therefore requires a proactive approach, incorporating predictive maintenance strategies and close collaboration with suppliers to ensure the continued availability of necessary parts and materials. 
SNCF Réseau, the French railway infrastructure manager, is responsible for maintaining, developing, and operating France's national rail network comprising approximately 30,000 kilometers of track and associated infrastructure. As a public service company and subsidiary of the SNCF Group, it plays a critical role in ensuring the safety, reliability, and efficiency of the French railway system while facilitating both passenger and freight transportation across the extensive rail corridors of the nation. 
SNCF Réseau uses durable but also aging equipment subject to the rigors of the environment. Customers, whether freight or passenger carriers, demand a high level of operational availability from their tracks. However, maintaining this availability becomes increasingly complex when equipment includes obsolete components. To optimize the management of obsolescence and shortages, SNCF Réseau is looking to strengthen its internal processes \citep{zolghadri2023micro}.

Within the literature, machine learning has a significant impact on many aspects of engineering, with the potential to enhance productivity and problem-solving capabilities, including decision-making \citep{tchuente2024methodological,culot2024artificial}. Consequently, machine learning has played a significant role in obsolescence management, specifically in the areas of obsolescence prediction \cite{trabelsi2021obsolescence}, forecasting obsolescence risk and product life-cycle \citep{jennings2016forecasting}, and classifying parts \citep{moon2022adaptive}. Machine learning, as defined by \citet{zhou2021machine}, is a set of algorithms that enables machines to learn and evolve behaviors based on data. Another related aspect for consideration is the shortage of data; not all companies deal with large amounts of available data, which is one of the biggest obstacles to tackle when trying to solve industrial problems using machine learning methods \citep{hubauer2013analysis,jess2015overcoming,libes2015considerations}.

Multiple machine learning-based models have been proposed for obsolescence forecasting \citep{jennings2016forecasting,grichi2017random,trabelsi2021obsolescence,liu2022obsolescence}, which rarely reach the ninetieth percentile in terms of accuracy, with most not even coming close. This highlights the critical need for improved forecasting techniques to address the challenges of component obsolescence, as noted by \citep{rust2022literature}. According to \citet{moon2022adaptive}, the limited success of machine learning and deep learning methods in obsolescence forecasting can be attributed to the scarcity of available data. This includes issues such as incomplete datasets and difficulties accessing relevant information, as highlighted by \citet{zolghadri2023micro}. Furthermore, as \citep{moon2022forecasting} points out, the limitations of available data further constrain the effectiveness of forecasting efforts. Therefore, the need for addressing the constraint of data availability becomes critical for the success of any machine learning-based obsolescence forecasting model. This need for data is particularly pronounced within SNCF Réseau, which funded this research, significantly hindering their capacity to implement proactive obsolescence management.

This paper presents a framework for obsolescence forecasting that addresses data availability constraints through a two-step approach. First, it augments the dataset using a deep generative model to mitigate the lack of data. Then, it applies a novel semi-supervised learning algorithm that enables classical machine learning models to learn from the augmented dataset and predict the state of parts.

To determine the most suitable generative model for obsolescence data generation, various state-of-the-art approaches are empirically compared. The forecasting model is then trained on the augmented dataset using the proposed semi-supervised learning algorithm, which leverages both labeled and unlabeled data to improve prediction accuracy. Since semi-supervised learning relies on inferring labels from the distribution of unlabeled data, high-quality synthetic data is crucial for achieving optimal performance \citep{cholaquidis2020semi}. This further underscores the importance of selecting the most effective generative model at generating obsolescence data. The proposed framework is tested and applied to two use case datasets to verify its validity.

The novelty of this work lies first in the proposed framework for obsolescence forecasting that addresses the data scarcity constraint that all machine learning-based models for obsolescence forecasting face. Furthermore, a novel semi-supervised learning algorithm is proposed to ensure that classical machine learning-based models for obsolescence forecasting are able to learn from the augmented dataset.

The remainder of this paper is segmented into the following sections: the state-of-the-art approaches for obsolescence forecasting are discussed in Section \ref{sec:literature_review}. Then, the proposed framework, including the proposed learning algorithm, is presented in Section \ref{sec:proposed_framework}. Continuing on, the evaluation methodology that has been used for testing the proposed framework is presented in Section \ref{sec:evaluation_methodology}. The different datasets used within this study are evaluated and discussed in Section \ref{sec:datasets}. Afterward, the experimentation results as well as the analysis are presented in Section \ref{sec:experiments}. Finally, some conclusions, discussion points, and future work are presented in Section \ref{sec:conclusion}. For simplicity and ease of reading, additional theoretical formulations are provided in appendix \ref{apdx:objective_function_formulation}, \ref{apdx:computational_cost}, and \ref{apdx:convergence_properties}. The details of the various experimentation methods can be found in \ref{apdx:evaluation_methodology_details}. Finally, details regarding the implementations are included in Appendix \ref{apdx:further_experimental_results}.

To ensure scientific replicability, the code and datasets used within this paper have been made publicly available on GitHub\footnote{https://github.com/Inars/Obsolescence-Forecasting-and-Data-Augmentation-Using-Deep-Generative-Models}.
\section{Literature Review}\label{sec:literature_review}
Traditionally, obsolescence forecasting relied on methods grounded in product life cycle models, often involving the analysis of characteristics like sales trends, pricing, and component usage to estimate the life stage of a product \citep{jennings2016forecasting}. Techniques such as fitting Gaussian trend curves to sales data were employed to predict zones of obsolescence \citep{jennings2016forecasting}. However, these methods often require subjective inputs, laborious data gathering, and struggle with the scale and complexity of modern systems involving thousands of components \citep{jennings2016forecasting}. The need for more automated, accurate, and scalable forecasting solutions has driven the adoption of data-driven approaches, particularly those based on machine learning.

The application of supervised machine learning techniques marked a significant step forward. \citet{jennings2016forecasting} provided early evidence of the potential of machine learning, demonstrating that algorithms such as Random Forests (RF), Artificial Neural Networks (ANN), and Support Vector Machines (SVM) could classify electronic components (specifically, cell phones scraped from the GSM Arena website) as active or obsolete with high accuracy and predict obsolescence dates within months \citep{jennings2016forecasting}. This work argued for the superiority of supervised learning over unsupervised or self-supervised methods for this task, citing the need for known component states (active/obsolete labels) in the training data and the difficulty in ensuring that clusters derived from unsupervised methods would align meaningfully with obsolescence status.

Subsequent research appeared to consolidate around the Random Forest algorithm as a particularly effective tool. \citet{grichi2017random} specifically advocated for RF, which were first introduced by \citet{breiman2001random}, highlighting its inherent resistance to variance and bias, applying it to the same GSM Arena dataset used by \citep{jennings2016forecasting}. The same author then proceeded to later combine a meta-heuristic genetic algorithm \citep{grichi2018optimization} and then a meta-heuristic particle swarm optimization algorithm \citep{grichi2018new} during training in order to increase the predictive efficacy of the Random Forest model.

Further reinforcement came from \citet{trabelsi2021obsolescence}, who tested multiple feature selection methods from the three feature selection approaches, i.e., filter, wrapper, and embedded, on five machine learning algorithms, namely RF, ANN, SVM, k-Nearest Neighbors (KNN), and naive Bayes. The author further amplified the conclusion presented by \citet{grichi2017random} and \citet{jennings2016forecasting}, which is that the RF model provided the highest precision and therefore the best model to use for predicting the risk of obsolescence. The authors have also scraped the GSM Arena website for technical data.

\citet{liu2022obsolescence} states that the amount of available data is limited and therefore affects the forecasting accuracy of obsolescence. They propose an obsolescence forecasting method which consists of two stages. In the first stage, the ELECTRE I method identifies key features to enhance prediction efficiency and accuracy of the model applied to the dataset obtained from \citet{jennings2016forecasting}. In the second stage, an improved radial basis function neural network is employed, incorporating a combination of information gain and ratio for data weighting, an enhanced particle swarm optimization algorithm for optimizing cluster centroids, and an improved gradient descent method for determining network weights. These improvements lead to better clustering, faster convergence, and higher prediction accuracy.

While the initial applications of supervised machine learning, particularly RF, demonstrated promising accuracy in classifying and predicting component obsolescence based on available datasets, a critical evaluation reveals several significant limitations that challenge the robustness and generalizability of these findings. These limitations stem from fundamental issues related to data availability and representativeness, as well as inherent constraints of the chosen algorithms and evaluation methodologies. Table \ref{tab:sota_summary} provides an overview and critique of the studies.

\begin{table*}
    \centering
    \begin{tabularx}{\textwidth}{@{} L L L L L L @{}}
    \toprule
    % Header Row (Row 1)
    \textbf{Reference} & \textbf{Core Method(s)} & \textbf{Dataset(s)} & \textbf{Contribution} & \textbf{Acknowledged Limitations} & \textbf{Identified Implicit Limitations \& Weaknesses} \\
    \midrule
    \citet{jennings2016forecasting} & RF, ANN, SVM & GSM Arena & Comparative study of algorithms & (Implicit) Scalability issues with traditional methods & Dataset dependency (GSM Arena); Limited generalizability; Assumes sufficient labeled data for supervised learning. \\
    \midrule
    \citet{grichi2017random} & RF & GSM Arena & Focus on RF due to bias/variance resistance & (Implicit) Potential for improvement via optimization & Dataset dependency; Limited generalizability; Data scarcity not explicitly addressed. \\
    \midrule
    \citet{grichi2018optimization} \& \citet{grichi2018new} & RF + GA; RF + PSO & GSM Arena & Metaheuristic optimization (GA, PSO) for RF parameters \& feature selection & Performance reduction due to irrelevant features & Optimization improves performance on existing data, does not solve scarcity; Risk of overfitting optimization to specific dataset; Dataset dependency persists. \\
    \midrule
    \citet{trabelsi2021obsolescence} & RF, ANN, SVM, kNN, NB + Feature Selection & GSM Arena & Comparison of Filter, Wrapper, Embedded feature selection methods & Need for relevant feature identification & Effectiveness depends on data representativeness; Does not solve scarcity; Dataset dependency persists; Wrapper/Embedded risk overfitting. \\
    \midrule
    \citet{liu2022obsolescence} & ELECTRE I + Improved RBFNN (PSO, GD variants) & GSM Arena & Two-stage approach: Feature identification (ELECTRE I) + Optimized RBFNN (weighting, PSO centroids, GD weights)  & Limited available data affects accuracy & Optimization still constrained by limited input data; Does not increase data quantity/diversity; Inherits data limitations. \\
    \bottomrule
    \end{tabularx}
    \caption{Summary and critique of machine learning approaches in obsolescence forecasting}
    \label{tab:sota_summary}
\end{table*}

A recurring and perhaps the most critical limitation acknowledged within the literature is the scarcity of available data for training robust obsolescence forecasting models \citep{moon2022forecasting}.  \citet{liu2022obsolescence} explicitly state that the limited amount of available data directly impacts forecasting accuracy \citep{sierra2023deep}. This observation is echoed across multiple studies, which identify data scarcity as a crucial weakness hindering the broader success and practical application of machine learning and deep learning methods in this domain \citep{liu2022obsolescence}. The problem is fundamental: machine learning models, particularly complex ones capable of capturing intricate patterns, generally require substantial amounts of data to learn effectively and generalize to unseen instances \citep{goodfellow2016deep}. Insufficient data can lead to models that fail to capture the underlying dynamics, exhibit high variance, or overfit to the limited training examples \citep{goodfellow2016deep}.

While techniques like metaheuristic optimization for parameter tuning \citep{grichi2018new}, feature selection \citep{trabelsi2021obsolescence}, and the development of advanced model architectures like improved radial basis function neural networks \citep{liu2022obsolescence} can enhance performance on existing datasets, they primarily represent model-centric refinements. They optimize how the model processes the available information but do not fundamentally resolve the underlying issues of data scarcity, dataset bias, or the limitations of relying solely on labeled data \citep{nassef2023review}. These approaches tend to operate within the constraints of the data rather than alleviating them.

Addressing these persistent gaps requires a shift in perspective, moving towards methodologies that directly tackle the root causes of the data limitations observed in prior art. The dual contribution proposed herein represents a significant departure from previous approaches documented in the literature. Instead of focusing solely on optimizing algorithms to perform marginally better on limited and potentially biased data, this work tackles the fundamental problem of data scarcity head-on through data generation. Concurrently, it introduces a learning methodology designed explicitly to thrive in such data-augmented environments.
\section{Proposed Framework}\label{sec:proposed_framework}
\begin{longtable}{|>{\centering\arraybackslash$}p{0.2\textwidth}<{$} | p{0.7\textwidth}|}
\hline
\textbf{Symbol/Notation} & \textbf{Description} \\
\hline

\mathcal{X} & Source domain (feature space), $\mathcal{X} \subset \mathcal{H}$. \\ \hline
\mathcal{Y} & Target domain (output space), $\mathcal{Y} \subset \mathcal{H}$. \\ \hline
\mathcal{Z} & Latent vector space, $\mathcal{Z} \subset \mathbb{R}^q$. \\ \hline
L & Set of labeled instances $\{(l_i, y_i)\}$. \\ \hline
N_L & Number of labeled instances. \\ \hline
L' & Dataset $L$ after dimensionality reduction, $r_{\psi}(L)$. \\ \hline
U & Set of unlabeled (or generated) instances $\{(u_i, y'_i)\}$. \\ \hline
N_U & Number of unlabeled (or generated) instances. \\ \hline
U' & Unlabeled/generated dataset in the reduced dimension space. \\ \hline
\mathcal{C} & Set of class values $\{1, ..., N_{\mathcal{C}}\}$. \\ \hline
\mathcal{D} & The complete dataset, $\mathcal{D} = L \cup U$. \\ \hline
N_{\mathcal{D}} & Total number of instances in the dataset $\mathcal{D}$, $N_L + N_U$. \\ \hline
Z & Set of latent samples drawn from $p(\mathcal{Z})$. \\ \hline
r_{\psi} & Transformation function for dimensionality reduction, parameterized by $\psi$. \\ \hline
d_{\phi} & Learning function of the discriminator/classifier model parameterized by $\phi$. \\ \hline
g_{\theta} & Learning function representing a generative model parameterized by $\theta$. \\ \hline
n & Dimensionality of the original data space $\mathcal{X} \subset \mathbb{R}^n$. \\ \hline
m & Dimensionality of the latent space after reduction ($m < n$). \\ \hline

D & Kolmogorov-Smirnov $D$ statistic. \\ \hline
l_1 & Wasserstein distance. \\ \hline
s_{\text{pearson}} & Pearson correlation similarity score. \\ \hline
s_{\text{coverage}} & Range Coverage score. \\ \hline
\log L & Log-Likelihood. \\ \hline
s_{\text{aAUC}} & Average Area Under the ROC Curve score. \\ \hline
s_{\text{acc}} & Accuracy score; proportion of correct predictions. \\ \hline
s_{\text{F1}} & F1-score. \\ \hline
s_{\text{AUC}} & Area Under the ROC Curve. \\ \hline
\mathcal{L}_{\text{RMSE}} & Root Mean Squared Error. \\ \hline
H(\cdot) & Differential Entropy. \\ \hline
I(\cdot) & Mutual Information. \\ \hline
\mathcal{L}_{\text{info}} & Information Loss. \\ \hline
C & TOPSIS relative closeness. \\ \hline

\caption{Glossary of symbols, variables, and notation} \label{tab:glossary_framework_nomethods}
\end{longtable}

The proposed framework consists of three parts: a dimensionality reduction part that simplifies the data into lower dimensional representations, a generator that produces synthetic data, and a discriminator that classifies this data into predefined states (e.g., available, obsolete). The generator uses three models representing the three distinct approaches to data generation. The discriminator is a classical machine learning-based model that employs a novel semi-supervised learning algorithm to predict component states. This method aims to improve forecasting accuracy by generating realistic synthetic data and accurately predicting the state of parts based on both labeled and unlabeled data.

The three generative models chosen are the following:
\begin{enumerate}
    \item \textbf{Real NVP:} For normalizing flows, the Real NVP (Non-Volume Preserving) \citep{dinh2016density} algorithm was chosen since, to the best of the knowledge of the authors, no other normalizing flow-based algorithm that focuses on tabular data was proposed in the literature.
    \item \textbf{VAE:} The Tabular Variational Autoencoder (TVAE) \citep{xu2020synthesizing} was chosen for it was made with tabular data in mind.
    \item \textbf{GAN:} The Conditional Tabular GAN (CTGAN) \citep{xu2020synthesizing} is a type of generative model designed to generate synthetic tabular data,
\end{enumerate}

For the purposes of the study, the Random Forest model has been selected to play the role of the discriminator of the proposed framework. The reasoning behind this decision is that the Random Forest model has been heavily studied in the literature \citep{jennings2016forecasting,grichi2017random,trabelsi2021obsolescence} and is demonstrated to achieve state-of-the-art results on the task of obsolescence forecasting.
\subsection{Problem Formulation}\label{sec:problem_formulation}
The following general mathematical formulation is geared towards encompassing both deep generative models as well as semi-supervised models in the same framework. For simplicity, the focus will be on the classification task of tabular data, which is the focus of the experimental work reported in this paper. However, with minimal modifications, the same formulation holds in more general settings. Let the data handled by the models within this paper be defined as follows
\begin{itemize}
    \item $\mathcal{X},\mathcal{Y}\subset\mathcal{H}:=(\mathbb{R}^{n},\langle\cdot,\cdot\rangle)$ are the source and target domains respectively assumed to be real finite-dimensional vector spaces. Both $\mathcal{X}$ and $\mathcal{Y}$, without loss of generality, are proper subspaces of the Hilbert space $\mathcal{H}$.
    \item $\mathcal{C}\subset\mathbb{Z}^+$ is the set of class values, such that $\mathcal{C}\equiv\left\{1,...,N_{\mathcal{C}}\right\}$ contains $N_{\mathcal{C}}$ classes.
    \item $L=\left\{(l_i,y_i)\in\mathcal{X}\times\mathcal{C}|i=1,...,N_L\right\}$ is the set of labeled instances.
    \item $U=\left\{(u_i,y'_i)\in\mathcal{X}\times\mathcal{C}|i=1,...,N_U\right\}$ is the set of unlabeled instances where the class value is to be estimated from $\mathcal{C}$, initially set to $-1$. The labeled and unlabeled sets are disjoint $L\cap U=\emptyset$.
    \item $\mathcal{D}=L\cup U$ is the dataset where $N_{\mathcal{D}}=N_L+N_U$ is the number of instances.
\end{itemize}

\begin{figure}[ht]
\centering
\resizebox{0.2\textwidth}{!}{%
\begin{circuitikz}
\tikzstyle{every node}=[font=\LARGE]
\draw  (5.5,11.25) rectangle (5.5,11.25);
\draw  (1.25,12.75) rectangle (8,-1);
\draw  (1.75,10.25) rectangle (6,5.75);
\draw  (7.5,10.25) rectangle (6.25,5.75);
\draw  (6.25,4) rectangle (7.5,-0.5);
\draw  (6,4) rectangle (1.75,-0.5);
\node [font=\LARGE] at (2.5,12) {};
\node [font=\LARGE] at (2.5,12) {};
\node [font=\LARGE] at (4.5,12) {$D$};
\node [font=\LARGE] at (4.5,10.75) {$L$};
\node [font=\LARGE] at (4.25,4.5) {$U$};
\draw  (1.5,5.25) rectangle (7.75,-0.75);
\draw  (1.5,11.5) rectangle (7.75,5.5);
\node [font=\LARGE] at (3.9,9) {$\vdots$};
\node [font=\LARGE] at (3.9,8) {$l_i$};
\node [font=\LARGE] at (3.9,7) {$\vdots$};
\node [font=\LARGE] at (6.9,9) {$\vdots$};
\node [font=\LARGE] at (6.9,8) {$y_i$};
\node [font=\LARGE] at (6.9,7) {$\vdots$};
\node [font=\LARGE] at (3.9,2.75) {$\vdots$};
\node [font=\LARGE] at (3.9,1.75) {$u_i$};
\node [font=\LARGE] at (3.9,0.75) {$\vdots$};
\node [font=\LARGE] at (6.9,2.75) {$\vdots$};
\node [font=\LARGE] at (6.9,1.75) {$y'_i$};
\node [font=\LARGE] at (6.9,0.75) {$\vdots$};
\draw [short] (2,1.75) -- (3.5,1.75);
\draw [short] (4.25,1.75) -- (5.75,1.75);
\draw [short] (2,8) -- (3.5,8);
\draw [short] (4.25,8) -- (5.75,8);
\end{circuitikz}
}%
\caption{Visualization of the dataset $D$ with its various components.}
\label{fig:dataset}
\end{figure}
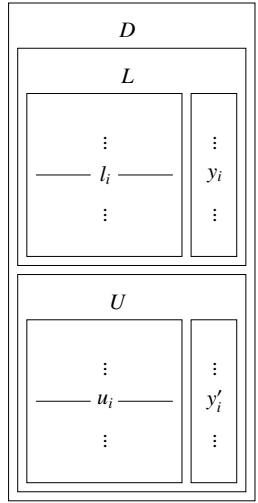

The dataset $\mathcal{D}$ as well as the sets $L$ and $U$ are shown visually in Figure \ref{fig:dataset}. The complete set of data $\mathcal{D}$, denoted as dataset, encompasses two subsets: a subset of non-generated data with known labels $L$ and a subset of generated data with unknown labels $U$. The subset $L$ is formed by the tuple $(l_i,a_i)\in\mathcal{X}\times\mathcal{C}$ such that $l_i$ is the feature vector and $a_i$ is the class label value. The subset $U$ is formed by the tuple $(u_i,b_i)\in\mathcal{X}\times\mathcal{C}$ such that $u_i$ is the generated feature vector and $b_i$ is the predicted class label value. The vector spaces $\mathcal{X}$ and $\mathcal{C}$ denote the feature space and the class value space respectively, meaning the spaces within which all possible features and class values are.

An ill-posed inverse problem as described by \citet{duff2024regularising}, is a problem that is solved by finding the best solution for which the forward analysis matches the desired output. Thus, the ill-posed inverse problem of predicting the state of components and systems is resolved using a parametric model denoted as a function $f$ that estimates the target given a set of data points and is governed by a set of parameters $\theta$, in other words $f_{\theta}:\mathcal{X}\rightarrow\mathcal{Y}$. Typically, solving ill-posed problems involves adding prior information, often through regularization, i.e., a term for the formulation of the problem to stabilize the solution, in a variational regularization framework, i.e., defining an objective function that adheres to the regularization. This is reflected by the optimization problem described by the following objective function:
\begin{equation}\label{eq:general_objective_function}
    \min_f\mathcal{L}(f_{\theta}(x_i),y_i)+\lambda\mathcal{R}(x_i),
\end{equation}
such that $\mathcal{L}:\mathcal{X}\times\mathcal{Y}\rightarrow[0,\infty]$ is a measurement function between the result obtained from applying the function $f_{\theta}$ representing the model to the input $x_i\in\mathcal{X}$ and the output $y_i\in\mathcal{Y}$, where $i=1,...,n$. The second term consists of $\mathcal{R}:\mathcal{X}\rightarrow[0,\infty]$ describing the regularization term which is small when the desired property is fulfilled, and the fixed regularization parameter $\lambda$ conditioned to be $\lambda\geq0$ which controls the impact between the data similarity term and the regularization term. Generally, the regularization term is done in the weight space in the form of L$2$ normalization introduced by \citet{srinivas2015deep} rendering $\mathcal{R}(x_i)=||x_i||_2^2$, or applying the dropout method introduced by \citet{srivastava2014dropout}.
\subsection{Framework Formalization}\label{sec:framework_formalization}
The proposed hybrid framework aims to train three models consecutively: an dimensionality reduction model that compresses the input data while preserving essential information, a generative model that captures the data distribution, and a discriminative model that estimates the probability of a sample belonging to a class $c_i\in\mathcal{C}$. For details regarding the formulation of the objective function of the framework, please refer to \ref{apdx:objective_function_formulation}. Additionally, details regarding the computational cost as well as the convergence of the framework can be found in \ref{apdx:computational_cost} and \ref{apdx:convergence_properties}, respectively.

To improve the effectiveness of the generative model and reduce noise in high-dimensional data, an initial dimensionality reduction step is introduced. Given an input set of labeled instances $L$, an invertible transformation function $r_{\psi}:\mathbb{R}^n\rightarrow\mathbb{R}^m$ parameterized by $\psi$ and conditioned by $m<n$, is applied to map the high-dimensional data into a lower-dimensional latent space while preserving critical structural information. The function $r_{\psi}$ is chosen to be invertible to allow for the reconstruction of generated data during inference. The reduced dataset, $L'$, serves as the input for both the generative and discriminative models.

Given a real finite dimensional latent vector space $\mathcal{Z}\subset\mathbb{R}^m$ with a tractable distribution, i.e., a distribution from which probabilities and samples can be computed, $p(\mathcal{Z})$, let $g_{\theta}:\mathcal{Z}\rightarrow\mathbb{R}^n$ be the learning function representing a generative model parameterized by $\theta$. Additionally, let $d_{\phi}:\mathcal{X}\cup\mathcal{Z}\rightarrow\mathcal{C}$ be the learning function of the discriminator model that is trained in the semi-supervised setting. The algorithm of the proposed framework is described in Algorithm \ref{alg:proposed_framework}. The pseudocode describes a hybrid framework designed to label an initially unlabeled dataset by using a two-step iterative process. The framework combines a generative model, $g_{\theta}$, and a discriminator model, $d_{\phi}$, to assign labels to data points in a semi-supervised way.

\begin{algorithm}
\caption{Pseudocode of the Proposed Framework}\label{alg:proposed_framework}
\SetKwInOut{Input}{input}\SetKwInOut{Output}{output}
\Input{$L$: Set of labeled set}
\Output{$\mathcal{D}$: Dataset with all $y'_i$ values are assigned}
Optimize $\psi$ by training $r_{\psi}(L)$\;
$L' = r_{\psi}(L)$\;
$Z\gets\emptyset$\;
\For{$i\leftarrow 0$ \KwTo $N_L-1$}{
$Z\gets Z\cup z_i\sim\mathcal{N}(\mu,\Sigma)$
}
Optimize $\theta$ by training $g_{\theta}(Z,L')$ following objective from Equation \ref{eq:objective_function}\;
$Z\gets\emptyset$\;
$Y\gets\emptyset$\;
\For{$i\leftarrow 0$ \KwTo $N_U-1$}{
$Z\gets Z\cup z_i\sim\mathcal{N}(\mu,\Sigma)$\;
$Y\gets Y\cup-1$
}
$U'\gets (g_{\theta}(Z),Y)$\;
Optimize $\phi$ through semi-supervised training $d_{\phi}(L',U')$ following objective from Equation \ref{eq:objective_function}\;
\For{$i\leftarrow 0$ \KwTo $N_U-1$}{
$(u_i,y'_i)\in U\gets(u_i,d_{\phi}(u_i))$
}
$U = r_{\psi}^{-1}(U')$\;
$\mathcal{D}\gets L\cup U$\;
\KwRet{$\mathcal{D}$}
\end{algorithm}

The algorithm starts by taking a labeled dataset, $L$, as input. To enhance the quality of learning, an initial dimensionality reduction model $r_{\psi}$ is learned and is applied to transform the high-dimensional data into a lower-dimensional representation, $L'$, using the function $r_{\psi}$. Next, an empty set, $Z$, is initialized to store latent samples. In the first stage, the algorithm builds $Z$ by drawing samples from a normal distribution $\mathcal{N}(\mu,\Sigma)$ for each labeled sample in $L'$, leveraging the maximum-entropy, isotropy, and reparameterization-friendly properties of the Gaussian distribution that help ensure stable and efficient training. These latent  vectors, along with the transformed labeled data $L'$, are used in combination with the labeled data to train the generative model $g_{\theta}$. Training $g_{\theta}$ involves optimizing its parameters according to the objective function in Equation \ref{eq:general_objective_function}.

Next, the algorithm prepares the unlabeled dataset $U'$. The set $Z$ is reset, and a placeholder label set $Y$ is initialized, with each unlabeled sample assigned a temporary label of $-1$. Using $g_{\theta}$, features are generated for $Z$ and paired with the placeholder labels in $Y$ to form the unlabeled dataset $U'$. This process essentially augments the unlabeled data with features generated by $g_{\theta}$.

The second stage involves training the discriminator model, $d_{\phi}$, using both the labeled dataset $L'$ and the generated unlabeled dataset $U'$. Like $g_{\theta}$, $d_{\phi}$ is optimized based on an objective function, allowing it to learn a decision boundary that can distinguish between the binary classes, i.e., available or obsolete. The novel semi-supervised learning algorithm is further explained in Section \ref{sec:learning_algorithm}.

Finally, once $d_{\phi}$ is trained, it is used to predict labels $y_i'$ for each unlabeled instance $u_i\in U'$. This step labels the previously unlabeled data, assigning each point a label based on the predictions of $d_{\phi}$. Afterwards, the unlabeled data is expanded to $U$ using the inverse transformation $r_{\psi}^{-1}$.

In the end, the complete $\mathcal{D}$ is then constructed by combining the original labeled data $L$ with the newly labeled instances in $U$. The algorithm returns $\mathcal{D}$, the fully labeled dataset, which can then be used for further analysis.

The proposed framework is possible due to two assumptions: i) the information preservation and ii) the invertibility. The first assumption highlights the importance of preserving as much information within the lower dimensions with the intention of reverting the process and reconstructing the original dimensions using the aforementioned lower dimensions, which is what the second assumption describes. These two assumptions are put in place with the intention of reconstructing the generated data points in $U$ so that the generated data is human legible to be used afterward by the SNCF Réseau personnel.

The autoencoder has been selected to reduce the dimensionality of the input space $\mathcal{X}$. Since the autoencoder learns to reconstruct the input without requiring labeled data, it operates in an unsupervised learning setting, which is ideal for the purposes of this study. As shown by \citet{janakiramaiah2020reducing} and \citet{wang2016auto}, autoencoders can learn representations that differ from—and may be more effective than—those produced by classical methods such as principal component analysis and Isomap. This feature addresses the information preservation constraints of this study. Another reason for selecting an autoencoder for the dimensionality reduction task is its decoder, which can map the reduced representations back to the original space, thereby satisfying the second invertibility constraint. Finally, autoencoders have been shown to enhance the accuracy of classification models more effectively than other dimensionality reduction methods, as demonstrated by \citet{fournier2019empirical}, thus justifying their use as the initial step in this framework.
\subsection{Learning Algorithm}\label{sec:learning_algorithm}
The proposed learning algorithm proposed within this body of work falls under self-training, which is a type of semi-supervised learning algorithm that iteratively uses its own predictions on unlabeled data to improve the model \citep{amini2022self}. The proposed learning algorithm applies a semi-supervised learning approach by clustering both labeled and unlabeled data, training a classifier model on clusters with sufficient labeled data, and iteratively using predictions to label previously unseen samples, ultimately producing a refined classifier model trained on both original and newly labeled data. The pseudocode of the novel learning algorithm is shown in Algorithm \ref{alg:semi-supervised_algorithm}.

\begin{algorithm}[ht]
\caption{Pseudocode of the Proposed Semi-Supervised Learning Algorithm}\label{alg:semi-supervised_algorithm}
\KwIn{$L = \{(l_i, y_i)\}_{i=1}^{N_L}$: Set of labeled data \\
      $U = \{(u_j, y'_j)\}_{j=1}^{N_U}$: Set of unlabeled data}
\KwOut{$d_{\phi}$: Final trained classifier model}
$X_{\text{tot}}\gets\{l_i\}_{i=1}^{N_L}\cup\{u_j\}_{j=1}^{N_U}$\;
$a\gets[\underbrace{1,\ldots,1}_{N_L},\underbrace{0,\ldots,0}_{N_U}]$\;
$y_{\text{tot}}\gets[y_1,\ldots,y_{N_L},\underbrace{-1,\ldots,-1}_{N_U}]$\;
$X_{\text{tot}}^{\text{scaled}} \gets \frac{X_{\text{tot}}-X_{\text{tot}}^{\min}}{X_{\text{tot}}^{\max}-X_{\text{tot}}^{\min}}$\;
$\kappa\gets\left\lfloor\frac{N_L+N_U}{\alpha}\right\rfloor$\;
$b\gets\text{KMeans}(K,X_{\text{tot}}^{\text{scaled}})$\;
\ForEach{$k\in\{0,\ldots,\kappa-1\}$}{
    $L_k\gets\{i\mid b[i]=k\land a[i]=1\}$\;
    $U_k\gets\{j\mid b[j]=k\land a[j]=0\}$\;
    \If{$|L_k|>0\land|U_k|>0$}{
        $X_k\gets X_{\text{tot}}[L_k,:]$\;
        $y_k\gets y_{\text{tot}}[L_k]$\;
        \If{$|\text{unique}(y_k)|=1$}{
            $y_{\text{tot}}[U_k]\gets y_k[1]$
        }
        \Else{
            $d_{\phi,k}\gets\text{Classifier}(X_k,y_k)$ \;
            $\hat{y}_{U_k}\gets\argmax(d_{\phi,k}(X_{\text{tot}}[U_k,:]))$\;
            $y_{\text{tot}}[U_k]\gets\hat{y}_{U_k}$
        }
    }
}
$X_{\text{tot}}^{\text{final}}\gets X_{\text{tot}}[y_{\text{tot}}\geq0,:]$\;
$y_{\text{tot}}^{\text{final}}\gets y_{\text{tot}}[y_{\text{tot}}\geq0]$\;
$d_{\phi}\gets\text{Classifier}(X_{\text{tot}}^{\text{final}},y_{\text{tot}}^{\text{final}})$\;
\KwRet{$d_{\phi}$}
\end{algorithm}

The input consists of two sets: $L$ the set of labeled data with known labels, and $U$ the set of unlabeled data with placeholder labels $y'_i$ set to $-1$. These datasets are merged into a single feature matrix $X_{\text{tot}}$, accompanied by a binary indicator vector $a$ that marks which points are labeled ($a_i=1$) and which are not ($a_i=0$). The labels for the merged dataset are stored in the vector $y_{\text{tot}}$, where labeled instances retain their original class labels $y_i$, while unlabeled data points are initialized with $-1$.

To standardize the data, the features in $X_{\text{tot}}$ are scaled using min-max normalization. This ensures all features are in the range $[0,1]$, making them comparable in the clustering step. The number of clusters $\kappa$ is determined dynamically as $\kappa=\left\lfloor\frac{N_L+N_U}{\alpha}\right\rfloor$ where $\alpha$ is an adjustable parameter. A K-Means clustering algorithm partitions the scaled dataset into $\kappa$ clusters, and the cluster assignments are stored in the vector $b$, where each element indicates the cluster index for the corresponding data point.

For each cluster $k$, the algorithm identifies two subsets: $L_k$, the set of labeled data points within the cluster, and $U_k$, the set of unlabeled points. If both subsets contain data ($|L_k|>0$ and $|U_k|>0$), the algorithm extracts the corresponding features $X_k$ and labels $y_k$ from $L_k$. If all points in $L_k$ share the same label, that label is directly assigned to all points in $U_k$. Otherwise, a classifier $d_{\phi,k}$ is trained on $X_k$ and $y_k$, and the predicted labels for the points in $U_k$ are determined by selecting the class with the highest probability from the output of the model $d_{\phi,k}$. These predicted labels are then assigned to the corresponding entries in $y_{\text{tot}}$.

Once all clusters are processed, the algorithm filters out any remaining unlabeled data points, yielding the final feature matrix $X_{\text{tot}}^{\text{final}}$ and label vector $y_{\text{tot}}^{\text{final}}$. A new classifier model, denoted $d_{\phi}$, is trained on this filtered dataset, which includes both the original labeled data and the formerly unlabeled points that were assigned labels during the clustering process. The final output of the algorithm is this trained model, $d_{\phi}$, which can now be used to predict labels for new, unseen data in Algorithm \ref{alg:proposed_framework}.
\section{Evaluation Methodology}\label{sec:evaluation_methodology}
The effectiveness of the proposed framework is assessed using standard evaluation metrics to ensure a comprehensive evaluation of its performance. These metrics provide insights into various aspects of the behavior of the proposed framework, including its ability to improve generalization and make accurate predictions. Each section of the framework is tested using a separate set of metrics. Elaborate explanations of each of the following metrics as well as their implementation can be found in \ref{apdx:evaluation_methodology_details}.
\subsection{Dimensionality Reduction Evaluation}\label{sec:dimensionality_reduction_evaluation}
The information preservation and invertibility of the autoencoder model are evaluated using three metrics:
\begin{enumerate}
    \item Reconstruction test: The Root Mean Squared Error (RMSE) \citep{chai2014root} measures the average magnitude of the error between the reconstructed values $l'_i$ and the original values $l_i$, by computing the square root of the mean of the squared differences. A lower RMSE value indicates better reconstruction capability of the model, as it reflects smaller discrepancies between the reconstructed and original values.
    \item Information preservation metric: Mutual Information (MI) \citep{carrara2020estimation} quantifies the statistical dependence between two variables by comparing how much knowledge of one variable reduces the uncertainty in the other. If MI is high, then knowing one variable provides significant information about the other, indicating a strong dependency; if MI is low, then knowledge of one variable offers little additional insight into the other, implying near independence.
    \item Information loss metric: Information loss \citep{baez2011characterization} quantifies how much entropy---an information-theoretic measure of uncertainty---of the original data $X$ is lost in its lower-dimensional latent representation $Z$, computed as $H(X)-H(Z)$ (see Appendix \ref{apdx:evaluation_methodology_details}). A smaller value indicates that more of the original information (entropy) is preserved in the latent space. To assess the results, one compares these information losses across different latent dimensionalities or different models: lower information loss suggests a more efficient representation that retains significant variability and structure from $X$, whereas higher loss indicates that important data features may be discarded in the dimensionality reduction process.
\end{enumerate}

To determine the most optimal autoencoder architecture, the Technique for Order of Preference by Similarity to Ideal Solution (TOPSIS) \citep{chakraborty2022topsis} is used. TOPSIS is a multi-criteria decision analysis method used to rank alternatives---in this case, different latent dimensions---based on their relative closeness to an ideal solution. The ideal solution is the one that maximizes the benefit criteria, such as MI, while minimizing cost criteria, including $m$, $\mathcal{L}_{\text{RMSE}}$, and $\mathcal{L}_{\text{info}}$ (see Appendix \ref{apdx:evaluation_methodology_details}). Conversely, the anti-ideal solution is the one that minimizes the benefit criteria and maximizes the cost criteria. TOPSIS ranks alternatives by calculating their Euclidean distances to the ideal and anti-ideal solutions. A TOPSIS score closer to 1 indicates better performance (closer to the ideal best and farther from the ideal worst), while a score closer to 0 indicates worse performance.
\subsection{Deep Generative Models Evaluation}\label{sec:deep_generative_models_evaluation}
Three types of evaluations are taken into account to validate the generative section of the proposed hybrid framework: direct, indirect, and efficiency evaluation.

First, a direct evaluation through a series of statistical metrics is used to quantify the various aspects of the generated data. These types of metrics are applied on both the original dataset $L'$ and the generated dataset $U'$ for comparison. Four statistical metrics are used, including:
\begin{enumerate}
    \item Statistical test: The Kolmogorov-Smirnov (KS) test is used to compare the distributions of the original and generated data to determine if they are pulled from the same underlying distribution \citep{hodges1958significance}. 
    \item Distance metric: The Wasserstein distance is a metric used to quantify the difference between two probability distributions, i.e., the original and generated probability distribution. A small Wasserstein distance, i.e., close to $0$, between the original and generated data indicates that the generated data distribution is close to the original data distribution.
    \item Correlation metric: The Pearson correlation coefficient is the metric chosen to calculate the linear relationship between two dimensions,  \citep{berman2016chapter}. The lower the score, i.e, close to $0$, the closer the pairwise correlations of the original and generated data are.
    \item Coverage metric: The range coverage metric evaluates whether the range of values in a dimension taken from the generated data fully encompasses the range of values found in the dimension taken from the original data. The higher the score, i.e, close to $1$, the more the generated columns cover the original column ranges. 
\end{enumerate}

Second, an indirect evaluation is done where model-based metrics are obtained. This type of evaluation is done not on the datasets themselves, but rather on the models that utilize both the original and generated datasets. Two model-based metrics are used, including:
\begin{enumerate}
    \item Likelihood metric: The log-likelihood metric is estimated using a Gaussian Mixture Model (GMM). A GMM is a probabilistic model that represents a distribution as a combination of multiple Gaussian (normal) distributions \citep{taboga2017lectures}. A higher log-likelihood means the model explains the data better. 
    \item Detection metric: This discrimination metric assesses the quality of generated data by measuring how challenging it is to differentiate it from original data using trained models. Logistic Regression (LR) and Support Vector Machine (SVM) algorithms are employed to quantify this difficulty. Detailed explanations of the LR and SVM algorithms can be found in \citet{taboga2017lectures} and \citet{goodfellow2016deep}, respectively. The accuracy metric is used to evaluate the classification ability. 
    A score of $1.0$ indicates that the model fails to distinguish between synthetic and real data, while a score of $0.0$ signifies that the model can flawlessly differentiate between the two data types.
\end{enumerate}

The voting system is a method used to determine the best generative model based on various evaluation metrics by directly comparing the metric values assigned to each model. In this context, each evaluation metric represents a distinct criterion for assessing the models, such as correlation, likelihood, and others. For each metric, the model that achieves the highest value is selected as the winner for that specific metric. The overall best generative model is then determined by tallying the number of wins across all metrics; the model with the most wins is declared the best.
\subsection{Semi-Supervised Algorithm Evaluation}\label{sec:semi-supervised_algorithm_evaluation}
Three evaluation metrics are used for the semi-supervised learning algorithm to evaluate its efficiency. The metrics are applied to the discriminator using the proposed learning algorithm. These metrics include:
\begin{enumerate}
    \item Accuracy value: This measures how often the discriminator correctly classifies inputs based on their predicted probabilities and actual outcomes. A high accuracy value (close to $1.0$) means the discriminator correctly classify most inputs, while a low accuracy (close to $0.0$) indicates it frequently misclassify them.
    \item F$1$-score: This metric evaluates the balance between precision (positive predictive value) and recall (true positive rate), providing a single measure of the effectiveness of the discriminator, especially when dealing with imbalanced datasets \citep{goodfellow2016deep}. A high F1-score signifies that the discriminator effectively identifies positive instances while minimizing false positives and negatives, with a maximum value of $1.0$ (indicating perfect precision and recall) and a minimum value of $0.0$ (indicating no positive predictions or complete misclassification).
    \item ROC AUC: Receiver Operating Characteristic - Area Under the Curve (ROC AUC) is a single scalar value summarizing the ability of a classifier to distinguish between positive and negative classes across all possible thresholds. Higher ROC AUC values (approaching $1.0$) indicate a better performing model that discriminates well between classes, while lower values (approaching $0.5$) suggest poor performance similar to random guessing.
\end{enumerate}
\subsection{Framework Evaluation}\label{sec:framework_evaluation}
To evaluate the proposed framework as a whole, the machine learning efficiency metric is used. This metric assesses how effectively synthetic data enhances the generalization ability of downstream tasks trained on this data \citep{xu2023utility}. Using the accuracy metric, four machine learning models were trained on the generated data $U'$ and tested on the real data $L'$ to evaluate the effectiveness of using synthetic data for machine learning prediction tasks. The four models include: Decision Tree (DT) \citep{Shalev-Shwartz_Ben-David_2014}, LR \citep{taboga2017lectures}, Adaptive Boosting (AdaBoost) \citep{hastie2009multi}, and Multilayer Perceptron (MLP) \citep{goodfellow2016deep}.
\section{Datasets}\label{sec:datasets}
Two case studies have been taken into consideration to test the proposed framework. The first case study is a high-level study on system obsolescence, which is reflected by the GSM Arena \citep{jennings2016forecasting} dataset, which is a dataset about smartphone obsolescence. This dataset is used as a benchmarking dataset within the literature. The second case study is a low-level study on component obsolescence, which is reflected by the Arrow \citep{saad_2024_15017365} dataset, which is a dataset about an electronic semiconductor component, the Zener diode. The data collected on the Zener diode was taken from the Arrow Electronics website, an electronic components distributor. 
\begin{table}[ht!]
    \centering
    \begin{tabular}{|c|c|c|c|c|c|c|}
        \hline
        \textbf{Dataset} & \textbf{Gran} & $\boldsymbol{N_L}$ & $\boldsymbol{N_0}$ & $\boldsymbol{N_1}$ & $\boldsymbol{n}$ & \textbf{\textbackslash} \\ \hline
        GSM Arena & coarse & $2890$ & $266$ & $2624$ & $29$ & $91$ \\ \hline
        Arrow & fine & $11080$ & $3500$ & $7580$ & $16$ & $68$ \\ \hline
    \end{tabular}
    \caption{Use case dataset specifications}
    \label{tab:dataset}
\end{table}

The various specifications of the two datasets are represented in Table \ref{tab:dataset}. The column $N_0=\sum_{i=1}^{N_L}\boldsymbol{1}(y_i=0)$ and $N_1=\sum_{i=1}^{N_L}\boldsymbol{1}(y_i=1)$ are the total number of available and obsolete instances respectively, such that $\boldsymbol{1}(\cdot)$ is the characteristic function that returns $1$ if the condition is met; otherwise $0$. The \textit{Gran} column represents the granularity of the dataset, meaning whether the dataset is fine-grained (i.e. electronic components) or coarse-grained (i.e. smartphones). The backslash (\textbackslash) column represents the percentage proportion of obsolete instances within the datasets. 

The GSM Arena is a relatively small and imbalanced dataset with a total of $2890$ labeled instances, partitioned into $266$ available and $2624$ obsolete instances. The obsolete instances constitute $90\%$ of the entire dataset. This dataset is coarse-grained, containing $29$ discrete and continuous features to describe the data, e.g., dimensions, weight, and display resolution.

The Arrow dataset, on the other hand, is relatively larger and more balanced, with a total of $11080$ labeled instances, partitioned into $2500$ available and $7580$ obsolete instances. The obsolete instances make up $68\%$ of the dataset. This dataset is fine-grained, consisting of $16$ discrete and continuous features to describe the data, e.g., nominal voltage (V), maximum power dissipation (MW), and packaging.
\section{Experimentation}\label{sec:experiments}
All experiments were conducted on a local system equipped with an Intel Xeon W-11955M CPU with a base frequency of 2.60 GHz and a maximum frequency of 2.61 GHz. The experiments were conducted in a secluded environment with a preset seed of $42$ to ensure reproducibility.
\subsection{Dimensionality Reduction Experiments}\label{sec:dimensionality_reduction_experiments}
The experimental results of the autoencoder models on the Arrow and GSM Arena datasets are represented in Tables \ref{tab:autoencoder_results_arrow} and \ref{tab:autoencoder_results_phone}, respectively. These tables illustrate the trade-offs between the four key criteria used for model evaluation: latent dimension, RMSE, mutual information, and information loss. The highest TOPSIS score is highlighted in bold. Further details regarding the autoencoder optimization procedure can be found in \ref{apdx:further_experimental_results}.

\begin{table*}[ht!]
    \centering
    \begin{tabular}{|c|c|c|c|c|c|}
        \hline
        $\boldsymbol{m}$ & $\boldsymbol{\mathcal{L}_{\text{RMSE}}}$ & $\boldsymbol{H(Z)}$ & $\boldsymbol{I(X;Z)}$ & $\boldsymbol{\mathcal{L}_{\text{info}}}$ & $\boldsymbol{C}$ \\ \hline
1 & 0.0742 & -199.0902 & -16.3379 & 77.0617 & \textbf{0.7621} \\ \hline
2 & 0.2316 & -47.4091 & 4.7134 & 228.7429 & 0.3372 \\ \hline
3 & 0.1933 & -65.8722 & 5.1878 & 210.2798 & 0.3466 \\ \hline
4 & 0.1566 & -84.0283 & -3.2945 & 192.1236 & 0.4205 \\ \hline
5 & 0.1288 & -103.4797 & 0.0108 & 172.6723 & 0.4239 \\ \hline
6 & 0.1085 & -122.3968 & -9.4905 & 153.7552 & 0.5298 \\ \hline
7 & 0.1048 & -141.1002 & -17.7611 & 135.0518 & 0.6192 \\ \hline
8 & 0.1067 & -160.5276 & -17.5316 & 115.6244 & 0.6226 \\ \hline
9 & 0.0801 & -180.2652 & -16.9325 & 95.8867 & 0.6424 \\ \hline
10 & 0.0742 & -199.0902 & -16.3379 & 77.0617 & 0.6416 \\ \hline
11 & 0.0879 & -218.5995 & -27.3288 & 57.5524 & 0.6984 \\ \hline
12 & 0.0612 & -238.0257 & -24.9565 & 38.1262 & 0.6939 \\ \hline
13 & 0.0668 & -257.6802 & -19.6513 & 18.4718 & 0.6564 \\ \hline
14 & 0.0905 & -277.4273 & -17.2590 & -1.2753 & 0.6229 \\ \hline
15 & 0.0818 & -4.0189 & -22.3464 & 272.1331 & 0.4587 \\ \hline
    \end{tabular}
    \caption{Autoencoder experimentation results on the Arrow dataset with an entropy value of $H(X)\approx-276.1520$}
    \label{tab:autoencoder_results_arrow}
\end{table*}

The Arrow dataset, characterized by a highly negative input entropy ($H(X)\approx-276.15$), exhibits a clear preference for the smallest latent dimension ($m=1$), which achieves the highest TOPSIS score ($C=0.7621$). Despite its minimal latent dimension, $m=1$ achieves the second-lowest RMSE ($0.0742$), underscoring the ability of the model to balance compression and reconstruction fidelity. Notably, the mutual information ($I(X;Z)=-16.34$) is negative for $m=1$. However, the TOPSIS score prioritizes the combination of criteria: low RMSE, minimal dimensionality, and moderate information loss ($\mathcal{L}_{\text{info}}=77.06$) for $m=1$ outweigh its suboptimal mutual information. As $q$ increases, RMSE initially rises (e.g., $m=2$: $\mathcal{L}_{\text{RMSE}}=0.2316$), and while mutual information becomes positive for intermediate dimensions (e.g., $m=2,\dots,5$), these gains are offset by higher reconstruction errors and increased dimensionality, leading to lower TOPSIS scores. The negative trend in $H(Z)$ (from $-199.09$ at $m=1$ to $-277.43$ at $m=14$) reflects increasing latent space disorder, further inflating information loss and reducing TOPSIS efficacy.

\begin{table*}[ht!]
    \centering
    \begin{tabular}{|c|c|c|c|c|c|}
        \hline
        $\boldsymbol{m}$ & $\boldsymbol{\mathcal{L}_{\text{RMSE}}}$ & $\boldsymbol{H(Z)}$ & $\boldsymbol{I(X;Z)}$ & $\boldsymbol{\mathcal{L}_{\text{info}}}$ & $\boldsymbol{C}$ \\ \hline
1 & 0.1794 & 1.9834 & -6.0331 & 38.1227 & 0.5534 \\ \hline
2 & 0.1123 & 8.6358 & -28.3779 & 44.7750 & \textbf{0.7177} \\ \hline
3 & 0.6422 & 3.2081 & 9.4924 & 39.3474 & 0.3352 \\ \hline
4 & 0.5426 & 2.8963 & 10.4051 & 39.0355 & 0.3542 \\ \hline
5 & 0.4841 & 4.1051 & 7.4823 & 40.2444 & 0.3822 \\ \hline
6 & 0.3902 & 5.1866 & -3.5733 & 41.3259 & 0.4404 \\ \hline
7 & 0.3185 & 3.5467 & -8.6656 & 39.6860 & 0.4832 \\ \hline
8 & 0.3299 & 2.5281 & -9.0511 & 38.6673 & 0.4824 \\ \hline
9 & 0.2378 & 3.9581 & -18.3982 & 40.0974 & 0.5607 \\ \hline
10 & 0.1794 & 1.9834 & -6.0331 & 38.1227 & 0.5000 \\ \hline
11 & 0.1834 & -0.6316 & -14.7315 & 35.5076 & 0.5454 \\ \hline
12 & 0.2068 & -3.2749 & -24.1645 & 32.8643 & 0.6130 \\ \hline
13 & 0.1867 & -2.0599 & -25.2169 & 34.0793 & 0.6206 \\ \hline
14 & 0.1757 & -4.9324 & -25.2037 & 31.2068 & 0.6218 \\ \hline
15 & 0.1660 & 10.5777 & -25.9472 & 46.7170 & 0.6047 \\ \hline
16 & 0.1504 & 10.3481 & -23.7166 & 46.4873 & 0.5853 \\ \hline
17 & 0.1171 & 9.5889 & -28.0827 & 45.7282 & 0.6110 \\ \hline
18 & 0.1486 & 10.0595 & -24.5482 & 46.1988 & 0.5788 \\ \hline
19 & 0.1048 & 8.8880 & -28.8217 & 45.0272 & 0.6000 \\ \hline
20 & 0.0946 & 8.6358 & -28.3779 & 44.7750 & 0.5933 \\ \hline
21 & 0.1112 & 6.3308 & -46.2596 & 42.4701 & 0.6684 \\ \hline
22 & 0.1104 & 1.9456 & -29.7179 & 38.0849 & 0.5954 \\ \hline
23 & 0.0951 & -3.2856 & -28.4159 & 32.8537 & 0.5848 \\ \hline
24 & 0.0925 & 1.1808 & -51.6081 & 37.3201 & 0.6648 \\ \hline
25 & 0.0661 & -0.5564 & -56.1160 & 35.5828 & 0.6699 \\ \hline
26 & 0.1124 & -2.6862 & -44.8528 & 33.4531 & 0.6328 \\ \hline
27 & 0.0648 & -10.2406 & -48.4456 & 25.8986 & 0.6442 \\ \hline
28 & 0.0742 & -7.7615 & -50.1696 & 28.3777 & 0.6374 \\ \hline
    \end{tabular}
    \caption{Autoencoder experimentation results on the GSM Arena dataset with an entropy value of $H(X)\approx-36.1393$}
    \label{tab:autoencoder_results_phone}
\end{table*}

For the GSM Arena dataset ($H(X)\approx-36.14$), the optimal latent dimension is $m=2$ ($C=0.7177$), balancing RMSE ($0.1123$) and information loss ($\mathcal{L}_{\text{info}}=44.78$). Unlike the Arrow dataset, mutual information remains negative across most dimensions, with $m=2$ achieving $I(X;Z)=-28.38$. Despite this, $m=2$ outperforms higher-dimensional models (e.g., $m=20$: $\mathcal{L}_{\text{RMSE}}=0.0946$, $C=0.5933$) due to its superior RMSE-dimensionality trade-off. Notably, larger latent dimensions (e.g., $m=25,\dots,28$) yield lower RMSE values (e.g., $m=25$: $\mathcal{L}_{\text{RMSE}}=0.0661$) but suffer from elevated information loss and dimensionality penalties, reducing their TOPSIS scores. The entropy $H(Z)$ for $m=2$ ($8.64$) is significantly higher than $H(X)$, indicating substantial information loss, yet this is counterbalanced by its competitive RMSE and minimal $q$.

The experimentation underscores the importance of multi-criteria optimization in autoencoder design. While theoretical metrics like mutual information and entropy provide foundational insights, their practical estimation challenges necessitate robust composite metrics like TOPSIS. The chosen models ($m=1$ for Arrow, $m=2$ for GSM Arena) exemplify the Pareto-optimal balance between reconstruction accuracy, latent dimensionality, information preservation, and mutual information.
\subsection{Deep Generative Models Experiments}\label{sec:deep_generative_models_experiments}
Table \ref{tab:generative_results} compares the performance of the generative models across the datasets, using multiple evaluation metrics. The bolded values highlight the best performance for a specific metric on a particular dataset, indicating where a model outperforms the others in that aspect. The implementation of the different generative models is detailed in \ref{apdx:further_experimental_results}.

\begin{table*}[ht!]
    \centering
    \begin{tabular}{|c|c|c|c|c|c|c|}
        \hline
        \textbf{Generator} & \multicolumn{2}{|c|}{\textbf{CTGAN}}  & \multicolumn{2}{|c|}{\textbf{TVAE}} & \multicolumn{2}{|c|}{\textbf{Real NVP}} \\ \hline
        \textbf{Dataset} & \textbf{Arrow} & \textbf{GSM Arena} & \textbf{Arrow} & \textbf{GSM Arena}& \textbf{Arrow} & \textbf{GSM Arena} \\ \hline
        \textbf{$\boldsymbol{D}$ statistic} & 0.0731 & 0.1005 & \textbf{0.0543} & 0.0593 & 0.1616 & \textbf{0.0476} \\ \hline
        \textbf{$\boldsymbol{p}$-value} & $8.49\times10^{-47}$ & $1.54\times10^{-16}$ & $\boldsymbol{5.85\times10^{-26}}$ & $4.62\times10^{-08}$ & $3.81\times10^{-228}$ & $\boldsymbol{6.77\times10^{-06}}$ \\ \hline
        \textbf{$\boldsymbol{l_1}$} & 0.281 & 1.779 & \textbf{0.166} & 1.538 & 0.526 & \textbf{0.966} \\ \hline
        \textbf{$\boldsymbol{s_{\text{pearson}}}$} & \textbf{0.0} & 0.209 & \textbf{0.0} & \textbf{0.0211} & \textbf{0.0} & 0.0290 \\ \hline
        \textbf{$\boldsymbol{s_{\text{coverage}}}$} & 0.289 & 0.917 & \textbf{0.319} & 0.659 & 0.142 & \textbf{0.964} \\ \hline
        \textbf{$\boldsymbol{\log L}$} & -2.656 & \textbf{-5.663} & -2.439 & -4.652 & \textbf{-3.561} & -5.237 \\ \hline
        \textbf{LR: $\boldsymbol{s_{\text{aAUC}}}$} & 0.976 & 0.860 & 0.997 & 0.946 & \textbf{1.000} & \textbf{0.991} \\ \hline
        \textbf{SVM: $\boldsymbol{s_{\text{aAUC}}}$} & \textbf{1.000} & \textbf{0.995} & 0.998 & 0.986 & 0.890 & 0.987 \\ \hline
    \end{tabular}
    \caption{Experimentation results of the comparison of generative models on two use case dataset.}
    \label{tab:generative_results}
\end{table*}

The experimental results comparing CTGAN, TVAE, and Real NVP on the Arrow and GSM Arena datasets reveal nuanced trade-offs in synthetic data quality across evaluation metrics. All models produced statistically significant differences between real and synthetic data distributions, as evidenced by $p$-values below $\alpha=0.05$ for the KS $D$-statistic. However, none met the critical thresholds for distributional fidelity ($D<0.013$ for Arrow, $D<0.025$ for GSM Arena). TVAE achieved the lowest $D$-statistic on Arrow ($D=0.0543$), while Real NVP performed best on GSM Arena ($D=0.0476$), though both values exceeded their respective thresholds. This suggests persistent challenges in replicating the original data distributions, particularly for GSM Arena, where all models exhibited higher divergence.

The Wasserstein distance ($l_1$) highlighted dataset-specific performance variability. TVAE excelled on Arrow ($l_1=0.166$), indicating its synthetic data closely aligns with the original distribution, while Real NVP outperformed others on GSM Arena ($l_1=0.966$). For correlation preservation, all models achieved perfect correlation ($s_{\text{pearson}}=0.0$) on Arrow due to the optimal latent dimension being $1$, but discrepancies emerged on GSM Arena. Here, TVAE ($s_{\text{pearson}}=0.0211$) and Real NVP ($s_{\text{pearson}}=0.0290$) outperformed CTGAN ($s_{\text{pearson}}=0.209$), with TVAE demonstrating the strongest ability to replicate original correlations.

Here, CTGAN ($s_{\text{pearson}}=0.209$) outperformed TVAE ($s_{\text{pearson}}=0.0211$) and Real NVP ($s_{\text{pearson}}=0.0290$). Coverage scores ($s_{\text{coverage}}$) further underscored model strengths: TVAE led on Arrow ($0.319$), while Real NVP dominated GSM Arena ($0.964$), suggesting robust capture of column ranges.

Log-likelihood ($\log L$) results highlighted distinct strengths. On Arrow, Real NVP achieved the best (most negative) score ($\log L=-3.561$), suggesting its synthetic data aligns well with the original distribution. TVAE ($\log L=-2.439$) and CTGAN ($\log L=-2.656$) followed. On GSM Arena, CTGAN achieved the lowest $\log L$ (-5.663), outperforming TVAE (-4.652) and Real NVP (-5.237). Detection metrics ($s_{\text{aAUC}}$) showed Real NVP excelling with LR classifiers ($1.000$ on Arrow, $0.991$ on GSM Arena), implying synthetic data indistinguishable from real data. CTGAN dominated with SVM classifiers ($1.000$ on Arrow, $0.995$ on GSM Arena), though high $s_{\text{aAUC}}$ values risk reflecting limited diversity.

Finally, using the voting system method, Real NVP emerged as the top-performing generator with a total of $8$ votes, followed by TVAE with $6$ votes, and CTGAN with $4$ votes. These results suggest that Real NVP demonstrates the most consistent ability to generate high-quality synthetic data across different metrics and datasets, making it the most suitable model for generating systems and components data for state forecasting tasks.
\subsection{Semi-Supervised Algorithm Experiments}\label{sec:semi-supervised_algorithm_experiments}
The results of the experimentation for the random forest model derived from the proposed semi-supervised learning algorithm are shown in Figure \ref{tab:discrimination_results}. The results are compared with those of the state-of-the-art method reported by \citet{trabelsi2021obsolescence}. Bold values indicate significant results, highlighting the highest performance for each metric on each dataset for both approaches. For further information regarding the details of the implementation, please refer to \ref{apdx:further_experimental_results}.

\begin{table*}[ht!]
    \centering
    \begin{tabular}{|c|c|c|c|c|c|c|c|}
        \hline
        \textbf{Model} & \multicolumn{6}{|c|}{\textbf{Proposed Framework}} & \textbf{\citet{trabelsi2021obsolescence}} \\ \hline
        \textbf{Generator} & \multicolumn{2}{|c|}{\textbf{CTGAN}}  & \multicolumn{2}{|c|}{\textbf{TVAE}} & \multicolumn{2}{|c|}{\textbf{Real NVP}} & \textbf{None} \\ \hline
        \textbf{Dataset} & \textbf{Arrow} & \textbf{GSM Arena} & \textbf{Arrow} & \textbf{GSM Arena}& \textbf{Arrow} & \textbf{GSM Arena} & \textbf{GSM Arena} \\ \hline
        $\boldsymbol{s_{\textbf{acc}}}$ & 0.9612 & 0.9815 & 0.9617 & 0.9830 & \textbf{0.9679} & \textbf{0.9836} & 0.9136 \\ \hline
        $\boldsymbol{s_{\textbf{F}1}}$ & 0.9557 & 0.9340 & 0.9566 & 0.9398 & \textbf{0.9632} & \textbf{0.9426} & 0.9190 \\ \hline
        $\boldsymbol{s_{\textbf{AUC}}}$ & 0.9561 & 0.9303 & \textbf{0.9697} & 0.9390 & 0.9692 & \textbf{0.9407} & --- \\ \hline
    \end{tabular}
    \caption{Efficiency metric values of the proposed discrimination algorithm compared to the state-of-the-art}
    \label{tab:discrimination_results}
\end{table*}

Across both the Arrow and GSM Arena datasets, all variants of the proposed framework consistently outperform the baseline, which lacks synthetic data augmentation. Notably, on the GSM Arena dataset, the proposed framework achieves marked improvements in accuracy ($s_{\text{acc}}$), F$1$ score ($s_{\text{F}1}$), and AUC ($s_{\text{AUC}}$), with Real NVP achieving the highest scores in $s_{\text{acc}}$ (0.9836) and $s_{\text{F}1}$ (0.9426), and TVAE yielding the highest $s_{\text{AUC}}$ (0.9697) on Arrow. This suggests that the choice of generative model influences performance, with Real NVP generally exhibiting superior discriminative capability, particularly in balancing precision and recall (as reflected in $s_{\text{F}1}$). 

The lower $s_{\text{acc}}$ (0.9136) and $s_{\text{F}1}$ (0.9190) of the baseline on GSM Arena underscore the critical role of synthetic data augmentation in addressing data scarcity, which the semi-supervised framework mitigates by enriching the training distribution. The results also reveal dataset-specific trends: GSM Arena exhibits marginally higher $s_{\text{acc}}$ and $s_{\text{F}1}$ values compared to Arrow, possibly due to inherent differences in data complexity or feature distributions. Overall, the integration of generative models, particularly Real NVP, with the semi-supervised pipeline significantly enhances the robustness of the underlying random forest classifier, validating the utility of the proposed framework in improving discrimination tasks through synthetic data augmentation.
\subsection{Proposed Framework Experiments}\label{sec:proposed_framework_experiments}
Table \ref{tab:ml_efficiency} displays the accuracy of various machine learning algorithms that were trained exclusively on the generated data. These algorithms were subsequently evaluated on the original Arrow and GSM Arena datasets to evaluate the effectiveness of the generated data.

\begin{table*}[ht!]
    \centering
    \begin{tabular}{|c|c|c|c|c|c|c|}
        \hline
        \textbf{Dataset} & \multicolumn{3}{|c|}{\textbf{Arrow}} & \multicolumn{3}{|c|}{\textbf{GSM Arena}} \\ \hline
        \textbf{Generator} & \textbf{CTGAN}  & \textbf{TVAE} & \textbf{Real NVP} & \textbf{CTGAN}  & \textbf{TVAE} & \textbf{Real NVP} \\ \hline
        \textbf{AdaBoost} & 0.9282 & 0.9282 & 0.9282 & 0.9260 & 0.9266 & 0.9256 \\ \hline
        \textbf{DT} & 0.9696 & 0.9741 & 0.9761 & 0.9671 & 0.9678 & 0.9900 \\ \hline
        \textbf{LR} & 0.8566 & 0.8589 & 0.8294 & 0.8360 & 0.8221 & 0.8370 \\ \hline
        \textbf{MLP} & 0.8570 & 0.9016 & 0.8585 & 0.9467 & 0.9561 & 0.9657 \\ \hline
    \end{tabular}
    \caption{Machine Learning efficiency metric in accuracy for different models and synthetic data methods.}
    \label{tab:ml_efficiency}
\end{table*}

The proposed framework achieves consistent high accuracy across both Arrow ($0.856$ to $0.976$) and GSM Arena datasets ($0.822$ to $0.989$), validating its integrated design. Dimensionality reduction via autoencoders preserves discriminative features, evidenced by LR achieving $0.829$ to $0.859$ accuracy despite synthetic data complexities. The semi-supervised label discrimination step ensures robust class boundary retention, with DTs reaching near-optimal performance ($0.969$ to $0.989$), directly attributable to the novel semi-supervised labeling algorithm. MLP results demonstrate context-dependent generator efficacy: TVAE excels on Arrow data ($0.902$ accuracy) through variational latent space alignment, while Real NVP dominates GSM Arena specifications ($0.966$) via normalizing flow advantages in high-dimensional feature capture. AdaBoost exhibits minimal performance variation ($\Delta<0.003$), confirming synthetic data stability across generators. These findings collectively confirm the framework reliably produces machine-learning-ready synthetic data, with component synergy compensating for individual architectural limitations.
\section{Conclusion}\label{sec:conclusion}
This paper presents a novel framework for obsolescence forecasting designed to enhance the performance of classical machine learning models by addressing the critical challenge of data scarcity. The framework integrates two core innovations: (1) a deep generative data augmentation module that synthesizes high-quality tabular data to expand limited training datasets, and (2) a semi-supervised learning algorithm that enables classical classifiers to leverage both labeled and unlabeled data effectively. By decoupling the generative and discriminative components, the framework ensures compatibility with any supervised learning model, eliminating the dependency on large labeled datasets. Experimental validation on industrial use cases demonstrates that the framework significantly improves forecasting accuracy for Random Forest classifiers (achieving $96.8\%$ accuracy on component-level data and $98.4\%$ on system-level data), outperforming state-of-the-art methods by $5$ to $7\%$, reaching the theoretical limit. The comparative analysis of generative models (Real NVP, TVAE, CTGAN), while secondary to the framework’s primary contribution, provides practical insights for selecting data synthesis strategies in obsolescence contexts. This work bridges the gap between data scarcity and robust machine learning in industrial applications, offering a flexible, model-agnostic solution for proactive obsolescence management.

The model-agnostic design of the framework opens several promising avenues for future research. First, its integration with emerging foundation models for tabular data could enable few-shot forecasting in extreme low-data regimes. Second, extending the semi-supervised algorithm to incorporate physics-informed constraints (e.g., reliability curves, wear-out mechanisms) may enhance alignment with domain-specific obsolescence drivers. Third, deploying the framework in streaming data environments--—where components are continuously monitored--—could enable dynamic risk assessment through incremental learning. By generalizing the architecture of the framework, future work could unify obsolescence prediction with related tasks like lifetime estimation and spare parts optimization, establishing a comprehensive toolkit for lifecycle management in long-field-life systems.
\appendix
\section{Objective Function Formulation}\label{apdx:objective_function_formulation}
The general objective introduced in Equation \ref{eq:general_objective_function} of the framework can be seen as a composite of three components:
\begin{equation}
    min[\mathcal{L}_{\text{REC}}+\mathcal{L}_{GEN}+\mathcal{L}_{\text{CLS}}]+\lambda\mathcal{R}(\cdot),
\end{equation}
where $\mathcal{L}_{\text{REC}}$, $\mathcal{L}_{\text{GEN}}$, and $\mathcal{L}_{\text{GEN}}$ are losses associated with dimensionality reduction, data generation, and classification, respectively, and $\mathcal{R}(\cdot)$ is a regularization term. The terms are weighted so that the significance of each part is properly reflected.

The autoencoder reconstruction term can be written as
\begin{equation}\label{eq:rec_objective_function}
    \mathcal{L}_{\text{REC}}(\psi)=\frac{1}{N_{\text{D}}}\sum_{i=1}^{N_{\text{D}}}||l_i-r_{\psi}^{-1}(r_{\psi}(l_i))||_2^2,
\end{equation}
ensuring preservability of critical information and promoting invertibility assumptions.

Depending on the generative model used, $\mathcal{L}_{\text{GEN}}$ is expressed differently:
\begin{itemize}
    \item TVAE: Minimizes the evidence lower-bound loss which consists of the reconstruction loss and the Kullback-Leibler divergence \citep{xu2020synthesizing}: 
    \begin{equation}
        \mathcal{L}=\mathcal{L}_{\text{recon}}-\mathcal{L}_{\text{KL}},
    \end{equation}
    where $\mathcal{L}_{\text{recon}}=\mathbb{E}_{q(z_i,l_i)}[\log p(l_i,z_i)]$ represents the reconstruction error between the true posterior density $p(l_i,z_i)$ and its approximate posterior $q(z_i,l_i)$. The term $\mathcal{L}_{\text{KL}}=D_{\text{KL}}(q(z|l)||p(z))$ denotes the Kullback-Leibler divergence, where $q(z_i|x_i)$ is the latent distribution and $p(z_i)$ is the prior (target) distribution. For categorical data points, the TVAE uses cross-entropy loss instead of Mean Squared Error (MSE).
    \item CTGAN: Uses an adversarial min–max game \citep{xu2020synthesizing}: 
    \begin{equation}
        \mathcal{L}=\min_G\max_D\mathbb{E}_{l\sim P_{L'}}[\log D(l,c)]+\mathbb{E}_{z\sim\mathcal{N}(\boldsymbol{0},\boldsymbol{I}),c\sim P_{L'}}[\log1-D(G(z,c),c)],
    \end{equation}
    where $z\sim\mathcal{N}(\boldsymbol{0},\boldsymbol{I})$ is the latent variable $z$ is sampled from the prior distribution that is assumed multivariate Gaussian prior distribution $\mathcal{N}(\boldsymbol{0},\boldsymbol{I})$. The conditional vector of categorical features $c\sim P_{L'}$ is sampled from the empirical distribution of the categorical features in the real data $c\sim P_{L'}$, denoted as $P_{L'}$.
    \item Real NVP: uses affine transformations that are both invertible and have tractable Jacobians. As a result, the log-likelihood objective can be expressed as follows \citet{dinh2016density}:
    \begin{equation}
        \mathcal{L}=\frac{1}{N_L}\sum_{i=1}^{N_L}\left[\log p_{z_i}(f(l_i))+\log\left|\det\left(\frac{\partial f(l_i)}{l_i}\right)\right|^{-1}\right],
    \end{equation}
    where $\frac{\partial f(l_i)}{l_i}$ is the Jacobian matrix of the transformation $f$ of data point $l_i$ and $p_{z_i}(f(l_i))$ is the data point $l_i$ after it has been transformed into the latent variable $z_i$.
\end{itemize}

For Random Forest, a typical surrogate could be a cross-entropy-like measure aggregated over the predictions. Denoting $d_{\Phi}$ as the trained forest, the classification loss is:
\begin{equation}
    \mathcal{L}_{CLS}(\phi)=_\frac{1}{N_{\mathcal{D}}}\sum_i\sum_r\boldsymbol{1}(y_i=r)\log(p(d_{\phi}(l_i)=r)),
\end{equation}
where $r$ indexes classes. In practice, RF often optimize impurity or Gini index at each node, but as an equivalent aggregate, it can be framed in a maximum-likelihood style.

Regularization penalties help avoid overfitting in high-dimensional settings. In this paper, weight-decay (L$2$) norms on model parameters are used for autoencoder and generative networks, dropout for neural networks, and pruning for Random Forest trees.

Hence, the full objective can be conceptualized as:
\begin{equation}\label{eq:objective_function}
    J(\psi,\theta,\phi)=\min(\mathcal{L}_{\text{REC}}(\psi)+\alpha_1\mathcal{L}_{GEN}(\theta)+\alpha_2\mathcal{L}_{CLS}(\phi))+\lambda\mathcal{R}(\psi,\theta,\phi),
\end{equation}
where $\alpha_1,\alpha_2\geq0$ tune the relative importance of generative fidelity versus classification accuracy, and $\lambda\geq0$ controls the extent of regularization.
\section{Computational Cost}\label{apdx:computational_cost}
Denote $m$ as the reduced dimensionality. In the proposed framework, the following major steps dictate the computational complexity:
\begin{enumerate}
    \item Dimensionality Reduction (Autoencoder Training): An autoencoder typically consists of an encoder $r_{\psi}$ and a decoder $r_{\psi}^{-1}$, parameterized by $\psi$. Training is done via backpropagation, which, for each update, costs $\mathcal{O}(n\times h)$ to process one sample, where $h$ is proportional to the total number of hidden-layer parameters in the autoencoder. Over $e_1$ training epochs with batch size $b_1$, training cost is approximately:
    \begin{equation}
        \mathcal{C}_{\text{AE}}\approx\mathcal{O}\left(e_1\times\frac{N_{\mathcal{D}}}{b_1}\times m\times h\right).
    \end{equation}
    \item Generative Model Training (TVAE, CTGAN, or Real NVP): , one typically minimizes a distribution alignment loss, i.e., Kullback–Leibler divergence, Wasserstein-based loss, or adversarial min–max. Each training iteration involves forward- and backpropagation through the generative model and possibly a critic (in the case of GAN-like approaches). If $e_2$ is the number of epochs and $b2$ the batch size for the generative model, then each epoch typically costs $\mathcal{O}(b_2\times G(\theta))$ per iteration, where $G(\theta)$ encapsulates the cost of passing a batch through the generative model (and possibly a discriminator component). Overall, the generative training complexity is:
    \begin{equation}
        \mathcal{C}_{\text{GEN}}\approx\mathcal{O}\left(e_2\times\frac{N_{\mathcal{D}}}{b_2}\times G(\theta)\right),
    \end{equation}
    where $G(\theta)$ for specifically TVAE, CTGAN, and Real NVP is the following:
    \begin{itemize}
        \item For TVAE, the cost involves encoding the data into latent space, decoding it, and minimizing a variational evidence lower bound, which entails sampling and backpropagation: $\mathcal{C}_{TVAE}=\mathcal{O}\left(e_2\times\frac{N_{\mathcal{D}}}{b_2}\times m\times q\right)$, where $q$ is the latent dimension.
        \item For CTGAN, it involves adversarial training with both generator and discriminator networks. Each iteration of training has complexity: $\mathcal{C}_{CTGAN}=\mathcal{O}\left(e_2\times\frac{N_{\mathcal{D}}}{b_2}\times(m\cdot\theta_g+m\cdot\theta_d)\right)$, where $\theta_g$ and $\theta_d$ are the parameters of the generator and discriminator, respectively. 
        \item For Real NVP, the computational cost is dominated by the transformations that encode and decode the data using invertible functions: $\mathcal{C}_{TVAE}=\mathcal{O}\left(e_2\times\frac{N_{\mathcal{D}}}{b_2}\times m\times k\right)$, where $k$ is the number of transformations in the model.
    \end{itemize}
    \item Discriminative Classifier (Random Forest): Random Forest training on $N_{\mathcal{D}}$ samples in $m$-dimensional space typically takes average-case time $\mathcal{O}(\#\text{Trees}\times N_{\mathcal{D}}\times m\times\log(N_{\mathcal{D}}))$ because each tree grows to about $\mathcal{O}(\log(N_{\mathcal{D}})$ depth, and one needs to evaluate m features for splits at each node. Denote $t$ as the number of trees, the cost becomes:
    \begin{equation}
        \mathcal{C}_\text{RF}\approx\mathcal{O}(t\times N_{\mathcal{D}}\times m\times\log(N_{\mathcal{D}})).
    \end{equation}
    \item Semi-Supervised Learning Algorithm: The semi-supervised learning algorithm of the framework, requires partitioning $N_{\mathcal{D}}$ samples into $\kappa$ clusters, each K-Means iteration costing $\mathcal{O}(N_{\mathcal{D}}\times m\times\kappa)$. If $i$ is the number of K-Means iterations (until convergence), the cost is thus approximately:
    \begin{equation}
        \mathcal{C}_{KM}\approx\mathcal{O}(i\times N_{\mathcal{D}}\times m\times\kappa).
    \end{equation}
\end{enumerate}
Combining these (and omitting additive constants for simplicity), the overall computational cost can be summarized as a sum of dominant terms:
\begin{equation}
    \mathcal{C}\approx\mathcal{C}_{\text{AE}}+\mathcal{C}_{\text{GEN}}+\mathcal{C}_{\text{KM}}+\mathcal{C}_{\text{KM}}.
\end{equation}
Substituting dependencies:
\begin{equation}
    \mathcal{C}\approx\mathcal{O}\left(e_1\times\frac{N_{\mathcal{D}}}{b_1}\times m\times h\right)+\mathcal{O}\left(e_2\times\frac{N_{\mathcal{D}}}{b_2}\times G(\theta)\right)+\mathcal{O}(N_{\mathcal{D}}\times m\times\kappa)+\mathcal{O}(i\times N_{\mathcal{D}}\times m\times\kappa).
\end{equation}
Since $\lambda$, the scale of regularization, and hyperparameters like batch size or epochs only affect constant multipliers, the growth is dominated by $N_{\mathcal{D}}$, $m$, $(\log(N_{\mathcal{D}}))$, and the architecture-specific parameters ($h$ for the autoencoder and $G(\theta)$ for the generative models).
\section{Convergence Properties}\label{apdx:convergence_properties}
Convergence in this framework refers to stabilizing three main components: (i) the dimensionality reduction (autoencoder), (ii) the generative model, and (iii) the semi-supervised classification. The influence of each part on the overall convergence is discussed in what follows:
\begin{enumerate}
    \item Dimensionality Reduction Convergence: Autoencoders minimize the construction loss in Equation \ref{eq:rec_objective_function} to train parameters $\psi$. In practice, standard results from stochastic gradient descent and variants, e.g., Adam, ensure that the reconstruction error converges to a local minimum. Once converged, the reduced representations $L'=r_{\psi}(L)$ are assumed to contain the salient features of the original data for subsequent modeling, thereby accelerating and stabilizing downstream tasks.
    \item Generative Model Convergence: For TVAE and Real NVP, training typically involves maximizing the likelihood of observed data or equivalently minimizing a divergence measure, i.e., Kullback–Leibler or Jensen–Shannon, between the data distribution and the model distribution. Under mild conditions and sufficient network capacity, these models converge to a local optimum that approximates the data-generating distribution. For CTGAN, the adversarial training approach involves a two-player minimax game. The generator aims to fool the discriminator, while the discriminator tries to distinguish real from fake samples. Although GANs can face convergence challenges, e.g., mode collapse, modern training heuristics and proper hyperparameter settings commonly lead to stable convergence.
    \item Semi-Supervised Classification Convergence: Random Forest classifiers do not rely on gradient-based methods; each tree converges once its splitting criteria can no longer reduce impurity. The entire ensemble converges when there are no more significant gains from adding new splits or after reaching a maximum depth. In the self-training semi-supervised phase, the iterative labeling of unlabeled data may in principle cause oscillations if predictions are inconsistent. However, once a stable cluster assignment and label propagation have emerged (with consistent assigned labels), the final pass of Random Forest training converges in a deterministic sense.
\end{enumerate}
By sequentially training the autoencoder (stable reconstruction), then the generative model (modeling data in the latent space), and finally the semi-supervised classifier (assigning or refining labels), the framework manages to decompose a large, complex optimization problem into specialized subproblems. Empirically, this stepwise approach typically yields stable solutions: the autoencoder ensures reduced-dimensional signals are clearer for generative training, and the generative model injects additional synthetic variety that, once combined with a robust classifier, avoids getting stuck in degenerate solutions.
\section{Evaluation Methodology Details}\label{apdx:evaluation_methodology_details}
This appendix provides detailed explanations of the various evaluation methodologies. Each performance evaluation metric provides an understanding of the performance of the proposed model in different areas and provides a different piece of information for understanding the bigger picture.

The RMSE is defined as:
\begin{equation}\label{eq:rmse}
    \mathcal{L}_{\text{RMSE}}=\sqrt{\frac{1}{N_{\mathcal{D}}}\sum_{i=1}^{N_{\mathcal{D}}}(l_i-l'_i)^2},
\end{equation}
where $N_\mathcal{D}$ is the number of data points, $l_i$ represents the actual value for the $i$-th observation, and $l'_i$ is the reconstructed value for that observation. The squaring of errors penalizes larger deviations more than smaller ones, making RMSE particularly sensitive to large errors.

Differential entropy is a measure of the uncertainty associated with a continuous random variable. For a random variable $X\in\mathbb{R^{N_{\mathcal{D}}\times n}}$ with probability density function $p(x)$, the differential entropy $H(X)$ is defined as:
\begin{equation}
    H(X)=-\int p(x)\log p(x)dx
\end{equation}
In practice, $p(x)$, is often unknown, $H(X)$ must be estimated
from data. Two approaches are considered: the Kozachenko-Leonenko k-NN estimator and the GMM-based estimator, where the k-NN estimator is used for $n\leq15$, while GMM is preferred for $n>15$ due to the curse of dimensionality.

The Kozachenko-Leonenko k-NN entropy estimator, based on the Kozachenko-Leonenko formula, estimates the differential entropy as follows:
\begin{equation}
    H(X)\approx\psi(N_{\mathcal{D}})-\psi(k)+\log(c_n)+\frac{n}{N_{\mathcal{D}}}\sum_{i=1}^{N_{\mathcal{D}}}\log(\epsilon_i),
\end{equation}
where $\psi(\cdot)$ is the digamma function, $k$ is the number of nearest neighbors, $c_n$ is the volume of the unit ball in $n$ dimensions, $\epsilon_i$ is the distance from the $i$-th sample to its $k$-th nearest neighbor. The volume of the unit ball in $n$ dimensions is given by $c_n=\frac{\pi^{\frac{n}{2}}}{\Gamma(\frac{n}{2}+1)}$ such that $\Gamma(\cdot)$ is the gamma function.

The GMM-based entropy estimator fits a Gaussian Mixture Model to the data and approximates the differential entropy as:
\begin{equation}
    H(X)\approx-\frac{1}{N_{\mathcal{D}}}\sum_{i=1}^{N_{\mathcal{D}}}\log p_{\text{GMM}}(x_i),
\end{equation}
where $p_{\text{GMM}}(x_i)=\sum_{k=1}^K\pi_k\mathcal{N}(x;\mu_k,\Sigma_k)$ is the probability density of the $i$-th sample under the fitted GMM.

MI $I(X;Z)$ measures the amount of information obtained about one random variable through another random variable. For continuous random variables $X$ and $Z$, MI is defined as:
\begin{equation}
    I(X;Z)=\int p(x,z)\log\frac{p(x,z)}{p(x)p(z)}dxdz.
\end{equation}
Two approaches are considered for estimating MI: the Kraskov-Stögbauer-Grassberger (KSG) estimator and the GMM-based estimator, where the KSG estimator is used for joint dimensions $n+q<20$, while the GMM estimator is preferred for $n+q\geq20$, due to the curse of dimensionality.

The KSG estimator for MI $I(X;Z)$ between $X\in\mathbb{R^{N_{\mathcal{D}}\times n}}$ and $Z\in\mathbb{R^{N_{\mathcal{D}}\times q}}$ avoids direct density estimation. Let $\epsilon_i$ be the distance to the $k$-th neighbor in the joint space $XZ$, and $d_x^{(i)}$, $d_z^{(i)}$, the counts of points within $\epsilon_i$ in $X$ and $Z$, respectively. Then:
\begin{equation}
    I(X;Z)\approx\psi(k)+\psi(N_{\mathcal{D}})-\frac{1}{N_{\mathcal{D}}}\sum_{i=1}^{N_{\mathcal{D}}}\left[\psi(d_x^{(i)}+1)+\psi(d_z^{(i)}+1)\right].
\end{equation}

The GMM-based MI estimator fits GMMs to the joint distribution $p(X,Z)$ and the marginals $p(X)$ and $p(Z)$. The MI is then estimated as:
\begin{equation}
    I(X;Z)\approx\frac{1}{N_{\mathcal{D}}}\sum_{i=1}^{N_{\mathcal{D}}}\left[\log p_{\text{GMM}}(x_i)+\log p_{\text{GMM}}(z_i)-\log p_{\text{GMM}}(x_i,z_i)\right],
\end{equation}
where $p_{\text{GMM}}(x_i)$, $p_{\text{GMM}}(z_i)$ and $p_{\text{GMM}}(x_i,z_i)$ are the probability densities under the fitted GMMs.

The information loss $\mathcal{L}_{\text{info}}$ is defined as the reduction in entropy between the input $X$ and latent representation $Z$:
\begin{equation}
    \mathcal{L}_{\text{info}}=H(X)-H(Z),
\end{equation}
where $H(X)$ and $H(Z)$ are estimated using the methods above. This assumes the encoder is deterministic, so $H(Z)\leq H(X)$, with equality only if no information is lost.

The TOPSIS method ranks alternatives based on their proximity to an ideal solution (minimizing all criteria) and distance from an anti-ideal solution (maximizing all criteria). We formalize this mathematically for the autoencoder optimization problem, where criteria include latent dimension $q$, reconstruction error $\mathcal{L}_{\text{RMSE}}$, mutual information $I(X;Z)$, and information loss $\mathcal{L}_{\text{info}}$. Let $A=\{A_1,\dots,A_m\}$ be a set of $m$ alternatives and let $C=\{C_1,\dots,C_n\}$ be a set of $n$ criteria. Each alternative $A_i$ is evaluated with respect to each criterion $C_j$, resulting in a decision matrix $D$:
\begin{equation}
    D=\begin{bmatrix}
x_{11} & x_{12} & \dots & x_{1n} \\
x_{21} & x_{22} & \dots & x_{2n} \\
\vdots & \vdots & \ddots & \vdots \\
x_{m1} & x_{m2} & \dots & x_{mn} \\
\end{bmatrix}
\end{equation}
where $x_{ij}$ is the value of the $j$-th criterion for the $i$-th alternative. To ensure that all criteria are comparable, the decision matrix $D$ is normalized. The normalized value $r_{ij}$ is computed as:
\begin{equation}
    r_{ij}=\frac{x_{ij}}{\sqrt{\sum_{i=1}^mx_{ij}^2}}
\end{equation}
This step ensures that each criterion is scaled to a unitless value, allowing for fair comparison across criteria. Each criterion $C_j$ is assigned a weight $w_j$, such that:
\begin{equation}
    W=[w_1,\dots,w_n],\sum_{i=1}^nw_i=1.
\end{equation}
The weighted normalized decision matrix $V$ is $v_{ij}=w_i\cdot r_{ij}$. Define the ideal best $A^+$ (minimized criteria) and ideal worst $A^-$ (maximized criteria):
\begin{align}
    A^+&=[\min_iv_{i1},\dots,\min_iv_{in}],\\
    A^-&=[\max_iv_{i1},\dots,\max_iv_{in}].
\end{align}
The Euclidean distance of each alternative $A_i$ to the ideal solution is computed as:
\begin{equation}
    d^+_i=\sqrt{\sum_{j=1}^n(v_{ij}-A^+_j)^2}.
\end{equation}
Similarly, the Euclidean distance of each alternative $A_i$ to the anti-ideal solution is computed as:
\begin{equation}
    d^-_i=\sqrt{\sum_{j=1}^n(v_{ij}-A^-_j)^2}.
\end{equation}
The relative closeness $C_i$ of each alternative $A_i$ to the ideal solution is defined as:
\begin{equation}
    C_i=\frac{d_i^-}{d_i^++d^-_i}.
\end{equation}
The value $C_i$ lies in the interval $[0,1]$, where $C_i=1$ indicates that the alternative is the ideal solution and $C_i=0$ indicates that the alternative is the anti-ideal solution. The alternatives are ranked in descending order of their relative closeness $C_i$. The alternative with the highest $C_i$ is considered the best.

The null hypothesis states that the two samples, i.e., the original and generated data, are drawn from the same distribution. The KS test calculates the difference ($D$) statistic, i.e., the maximum difference between the two given empirical cumulative distribution functions, given by:
\begin{equation}
    D_{N_L,N_U}=\sup_x|F_{L,N_L}(x)-F_{U,N_U}(x)|,
\end{equation}
where $F_{L,N_L}$ and $F_{U,N_U}$ are empirical cumulative distribution functions of the original and generated sample respectively, and are defined as:
\begin{equation}
    F_N(x)=\frac{1}{N}\sum_{i=1}^N\boldsymbol{1}_{(-\infty,x]}(x_i),
\end{equation}
where $\boldsymbol{1}_{(-\inf,x]}(x_i)$ is the characteristic function returning $1$ if data point $x_i<x$ and $0$ otherwise for a fixed $x$, and $N$ is the total number of data points in a given sample. A larger $D$ value indicates a greater discrepancy between the distributions being compared. The null hypothesis is rejected at a predefined significance level $\alpha=0.05$ following the condition:
\begin{equation}
    D_{N_L,N_U}>c(\alpha)\sqrt{\frac{N_L+N_U}{N_LN_U}},
\end{equation}
where the critical value $c(\alpha)$ is defined as:
\begin{equation}\label{eq:critical_threshold}
    c(\alpha)=\sqrt{-\ln\left(\frac{\alpha}{2}\right)\cdot\frac{1}{2}}.
\end{equation}
The limiting cumulative distribution function ($p$-value) is also approximated using the following infinite sum \citep{smirnov1948table}:
\begin{equation}
    L(z)\approx1-2\sum_{i=1}^{\infty}(-1)^{i-1}e^{-i^2z^2},
\end{equation}
where $z=D_{N_L,N_U}\sqrt{\frac{N_LN_U}{N_L+N_U}}$. The $p$-value indicates the probability of observing a KS statistic as extreme as the one calculated from the data under the null hypothesis. The null hypothesis is rejected if the $p$-value follows the following condition $L(\alpha)\leq\alpha$ with a predefined significance level $\alpha=0.05$, indicating a significant difference between the original and generated data distributions.

The Wasserstein distance is calculated using the following Euclidean norm equation:
\begin{equation}                    l_1(p_r,p_{\theta})=\inf_{\gamma\in\Gamma(p_r,p_{\theta})}\int||x-y||_2d\gamma(x,y),
\end{equation}
where $\Gamma(p_r,p_{\theta})$ is the set of probability distributions on $\mathcal{X}\times\mathcal{Y}$ and $p_r$ and $p_{\theta}$ are the marginals on the first and second factors respectively such that for a given original data point $x$, $p_r(x)$ returns the probability of $p_r$ at position $x$ and for a given generated data point $y$, $p_{\theta}$ returns the probability of $p_{\theta}$ at position $y$. 

The Pearson correlation quantifies how strongly two variables are related and whether the relationship is positive, negative, or nonexistent. The Pearson correlation coefficient is given by the following equation 
\begin{equation}:
    r_{x^a,x^b\in\mathcal{X}}=\frac{\sum_{i=1}^N(x_i^a-\Bar{x}^a)(x_i^b-\Bar{x}^b)}{\sqrt{\sum_{i=1}^N(x_i^a-\Bar{x}^a)^2}\sqrt{\sum_{i=1}^N(x_i^b-\Bar{x}^b)}},
\end{equation}
where $N$ is the number of data points in the original or generated data sample, $\Bar{x}=\frac{1}{N}\sum_{i=1}^Nx_i$ is the sample mean, and $x_i^a$ and $x_i^b$ are the individual sample data points such that $a$ and $b$ represent the index of two different columns. In the case of this paper, $a=0$ and $b=1$. The following similarity score is returned after running the test on the original data and the generated data:
\begin{equation}
    s_{\text{pearson}}(L,U)=\frac{|r_{x^a,x^b\in U}-r_{x^a,x^b\in L}|}{2}.
\end{equation}

The range coverage metric score is calculated using the following equation:
\begin{equation}
    s_{\text{coverage}}\left(x^{i,L},x^{i,U}\right)=1-\Biggl[\max\left(\frac{\min(x^{i,U})-\min(x^{i,L})}{\max(x^{i,L})-\min(x^{i,L})},0\right)+\max\left(\frac{\max(x^{i,L})-\max(x^{i,U})}{\max(x^{i,L})-\min(x^{i,L})},0\right)\Biggr],
\end{equation}
where $x^{i,L}$ and $x^{i,U}$ represent the original and generated columns for dimension $i$. The final score is given by the mean score $\Bar{s}=\frac{1}{2}(s_{x^{0,L},x^{0,U}}+s_{x^{1,L},x^{1,U}})$ for columns of dimension $0$ and $1$.

Regarding the GMM, consider a family of mixtures of $K$ multivariate normal distributions parameterized by the set of parameters $\Theta=\{\alpha_1,...,\alpha_K,\theta_1,...,\theta_K\}$ where $\theta_i=\{\mu_i,\Sigma_i\}$ is the parameter of the $i$-th normal distribution $p_i(x|\theta_i)$, $\mu_i\in\mathbb{R}^d$ is the mean vector, and $\Sigma_i\in\mathbb{R}^d$ is the covariance matrix. The objective of the GMM is to estimate the parameters $\Theta$ that maximize the log-likelihood function:
\begin{equation}
    \log L(\Theta|X)=\log p(X|\Theta)=\sum_{j=1}^N\log\left(p(x_j|\Theta)\right),
\end{equation}
where $N$ is the total number of generated and original instances, $X\subset\mathcal{X}$ is a set of data points independent and identically distributed (i.i.d.) sampled from the original data space, $p(x_j,\Theta)=\sum_{i=1}^K\alpha_ip_i(x_j,\theta_i)$ is the mixture joint probability density function and $\alpha_i\in[0,1]$ is the mixing probability coefficient (weight) under the constraint $\sum_{i=1}^K\alpha_i=1$. The expectation-maximization algorithm is used to iteratively update the parameters of each normal distribution in the mixture, where the details of the algorithm steps are explained by \citet{taboga2017lectures}. Multiple GMMs are used to cover the entirety of the data space $\mathcal{X}$. The Bayesian information criterion is used to determine the optimal number of GMMs to be used, defined by the equation \citep{hastie2009elements}:
\begin{equation}
    s_{\text{BIC}}=-2\log L(X|\Theta)+|\Theta|\log(N_{\mathcal{D}}).
\end{equation}

The ROC curve is formed by plotting the cumulative distribution functions of true positive rate (TPR), also known as sensitivity or recall, against the false positive rate (FPR), which is defined as $1$ minus specificity, at various threshold settings, summarized by their equations \citep{junge2018roc}:
\begin{equation}\label{eq:tpr}
    s_{\text{TPR}}(\tau)=\frac{\sum_{i=1}^N\boldsymbol{1}(p(x_i)\geq\tau,y_i=1)}{\sum_{i=1}^N\boldsymbol{1}(y_i=1)}=\int_{-\infty}^{\tau}p(\mathcal{X})dx\text{ for }y=1,
\end{equation}
\begin{equation}
    s_{\text{FPR}}(\tau)=\frac{\sum_{i=1}^N\boldsymbol{1}(p(x_i)\geq\tau,y_i=0)}{\sum_{i=1}^N\boldsymbol{1}(y_i=0)}=\int_{-\infty}^{\tau}p(\mathcal{X})dx\text{ for }y=0,
\end{equation}
where $N$ is the total number of generated and original instances, $\tau=0.5$ is the decision threshold of a classification, $\boldsymbol{1}(\cdot)$ is the characteristic function, and $p(x_i)$ is the probability of instance $x_i$ belonging to class $y_i$. 

The AUC is a scalar value that quantifies the overall performance of the classifier across all possible discrimination thresholds. The AUC is defined as the area under the ROC curve \citep{calders2007efficient}, which can be expressed as:
\begin{align}
    s_{\text{AUC}}(\tau)&=P(p(\mathcal{X}_i)>p(\mathcal{X}_2|y_1=1,y_2=0))\notag\\&=\int_0^1s_{TPR}(\tau)d(s_{FPR}(\tau)).
\end{align}
The final score is based on the average AUC score given by
\begin{equation}
    s_{\text{aAUC}}(\tau)=1-(2\cdot\max(s_{AUC},\tau)-1).
\end{equation}

Accuracy is defined as the ratio of correct predictions to the total number of instances \citep{goodfellow2016deep}, which can be written as
\begin{equation}
    s_{\text{acc}}(\tau)=\frac{\sum_{i=1}^N\boldsymbol{1}[p(x_i)\geq\tau,y_i=1]+\boldsymbol{1}(p(x_i)<\tau,y_i=0)}{N},
\end{equation}
where $N$ is the total number of generated and original instances, $p(x_i)$ is the predicted probability for the $i$-th instance, $y_i$ is the actual outcome for the $i$-th instance, and $\boldsymbol{1}(\cdot)$ is the characteristic function which equals to $1$ if the conditions inside are true and $0$ otherwise.

Precision, also known as confidence, measures how many of the predicted positive cases are actually positive, and is given by
\begin{equation}
    s_{\text{prec}}(\tau)=\frac{\sum_{i=1}^N\boldsymbol{1}(p(x_i)\geq\tau,y_i=1)}{\sum_{i=1}^N\boldsymbol{1}(p(x_i)\geq\tau)}
\end{equation}
where $N$ is the total number of generated and original instances, $p(x_i)$ is the predicted probability for the $i$-th instance, $y_i$ is the actual outcome for the $i$-th instance, and $\boldsymbol{1}(\cdot)$ is the characteristic function which equals to $1.0$ if the conditions inside are true and $0.0$ otherwise.

The F$1$-score is the harmonic mean of precision and recall from Equation \ref{eq:tpr}, and is given by
\begin{equation}
    s_{\text{F}1}(\tau)=2\frac{s_{\text{prec}}(\tau)s_{\text{TPR}}(\tau)}{s_{\text{prec}}(\tau)+s_{\text{TPR}}(\tau)}.
\end{equation}
\section{Further Experimental Results}\label{apdx:further_experimental_results}
The methodology for determining the optimal autoencoder architecture across latent dimensions involves a systematic hyperparameter search combined with regularization and validation-based early stopping. Autoencoders are constructed with symmetrical encoder-decoder architectures, where the encoder reduces the input dimensionality through a series of hidden layers to a latent bottleneck, and the decoder reconstructs the input by mirroring the structure of the encoder. Key architectural hyperparameters include the number of hidden layers (fixed at $2$), node permutations per layer (swept from $64$ to $256$ in increments of $64$), and latent dimensions (ranging from $1$ to the input dimensionality in unit steps). Each dense layer incorporates $L2$ regularization ($\lambda=1\times10^{-4}$) to mitigate overfitting, with ReLU activations for hidden layers and linear activation for the output. The training protocol employs the Adam optimizer with MSE loss, a maximum of $500$ epochs, and early stopping triggered after $10$ epochs of non-improving validation loss ($20\%$ validation split). The MSE loss is defined as:
\begin{equation}
    \mathcal{L}_{\text{MSE}}=\frac{1}{0.2\cdot N_{\mathcal{D}}}\sum_{i=1}^{N_{\mathcal{D}}}||l_i-D(E(l_i))||_2^2.
\end{equation}
For each latent dimension, all node permutations ($4^2=16$ architectures) are trained, and the model with the lowest test MSE is selected as optimal. 

Post-training analysis computes per-feature reconstruction differences between untrained and trained models to quantify feature-specific learning efficacy, with latent representations and trained models archived for downstream tasks. The approach prioritizes architectural symmetry, exhaustive latent dimension exploration, and regularization to balance reconstruction fidelity with generalization, while full-batch training on standardized data ensures stable convergence across varying dataset scales. Model selection emphasizes both performance (minimum test MSE) and parsimony, favoring simpler node configurations when performance differences are statistically insignificant ($\alpha=0.05$ via paired t-test). This structured yet flexible framework enables reproducible identification of latent spaces that optimally trade off compression and reconstruction accuracy.

For the purposes of the experiments, CTGAN is configured with key hyperparameters designed to balance the training of its generator and discriminator. To achieve this, both components use a learning rate of $2\times10^{-4}$, which controls the step size for each update, ensuring neither component learns too quickly nor too slowly. Additionally, a small weight decay of $1\times10^{-6}$ is applied through the Adam optimizer; this decay slightly penalizes larger weight values, helping to prevent overfitting by encouraging the model to learn general patterns rather than noise. This leads to a regularization coefficient of $\lambda=2\times10^{-4}\cdot1\times10^{-6}$. The discriminator updates once for every generator update, matching the original CTGAN design by \citet{xu2020synthesizing} to maintain stable training. Additionally, log frequency sampling, which ensures that categorical variables are accurately represented, especially for rare categories, is ignored since the inputs are real valued. The pseudo-assembled critic mechanism groups $10$ samples at a time for discriminator evaluation, allowing the model to capture relationships between samples rather than evaluating each one independently; this approach enhances the discriminator's ability to identify subtle patterns across batches. The model is trained for $300$ epochs processing $500$ samples per batch, meaning it passes through the entire dataset 300 times, which is intended to provide sufficient time for convergence, where the generator and discriminator reach a balance without further significant improvement.

The TVAE model focuses on balancing reconstruction quality and latent space learning. It applies L$2$ regularization with a value of $\lambda=1\times10^{-5}$ to control overfitting and introduces a loss factor of $2$ to weight the reconstruction error more heavily. This helps the model maintain a meaningful latent space while ensuring that reconstructed data closely resembles the original input. TVAE processes $500$ samples per batch during training and runs for $300$ epochs.

The RealNVP architecture employs $6$ affine coupling layers with alternating bipartite masks to ensure triangular Jacobians and full feature interaction, each containing conditioner networks (hidden dimensions = 512) with ReLU-activated MLPs, while scale parameters are clamped (scale clamp = 1) via a $\tanh$ nonlinearity to stabilize exponential transformations \citep{dinh2016density,papamakarios2021normalizing}. Training utilizes a conservative learning rate $1\times10^{-5}$ and batch size of $128$ to mitigate gradient instability in likelihood estimation, complemented by reduced learning rate on plateau scheduling and early stopping to prevent overfitting. The masking strategy alternates between even and odd indices across layers, enabling universal approximation capabilities \citep{kobyzev2020normalizing}, while the isotropic Gaussian base distribution ensures tractable density estimation.

The semi-supervised learning algorithm code integrates a suite of hyperparameters and methodologies designed to address challenges in semi-supervised learning, emphasizing robustness, computational efficiency, and reproducibility. Central to the framework is the use of Isolation Forest for outlier detection, configured with $100$ estimators to balance computational cost and detection accuracy. The dynamic selection of maximum features ($30\%$ of input features) ensures diversity across trees while mitigating overfitting, and a contamination rate of $5\%$ reflects a conservative assumption about outlier prevalence. Parallel execution via CPU core utilization accelerates training. For classification, the Random Forest model employs $100$ fully grown trees (maximum depth is not set) to minimize bias, with fixed random seeds ensuring reproducibility across runs. The parallelization strategy of the model optimizes resource usage, critical for large-scale sets.

Clustering plays a pivotal role in the semi-supervised pipeline. The K-Means algorithm is tuned by evaluating cluster counts between 2 to 5 using silhouette scores, avoiding overclustering in smaller subsets. Key parameters include k-means++ initialization for stable centroid placement, a reduced maximum iteration count ($50$) with relaxed tolerance (tolerance$=1\times10^{-2}$) for early convergence, and the number of times the algorithm is run with different centroid seeds is set to $3$ to limit computational overhead. For subsets with fewer than $50$ samples, heuristic labeling replaces clustering: $1$D subsets use median-based thresholds, while $2$D data employs quadrant splits. Clusters guide label propagation, where unlabeled instances inherit predictions from cluster-specific RF trained on labeled subsets. This localizes decision boundaries, leveraging inherent data structures. Clusters with homogeneous labels propagate dominant classes directly, ensuring stability in low-diversity regions.

Data preprocessing and validation strategies further enhance robustness. A minmax normalization function standardizes features to $[0,1]$ prior to clustering, addressing the sensitivity of K-Means to feature scales. Outlier removal via Isolation Forest is applied iteratively to the augmented sets, improving data quality. Stratified $5$-fold cross-validation preserves class distributions during evaluation, reducing bias in performance metrics for imbalanced sets. Parallel hyperparameter searches for K-Means expedite cluster evaluation, while subsampling ($1,000$ instances) accelerates silhouette score calculations without sacrificing reliability. These optimizations balance accuracy and computational cost, particularly for larger sets.

Computational efficiency is prioritized through early stopping, subsampling, and parallelization. The relaxed convergence criteria of K-Means reduce iteration counts, while cached scalers avoid redundant preprocessing during cross-validation. Reproducibility is enforced via fixed random seeds across all stochastic processes, including data splitting, model initialization, and clustering. The deterministic pipeline ensures consistent results, essential for experimental validation. Adaptive fallbacks, such as heuristic labeling for small datasets, maintain functionality in edge cases where traditional clustering fails.

For the machine learning efficiency models, the Adaptive Boosting classifier employed default settings for its base estimator (Decision Tree with maximum depth $=1$) and learning rate ($1.0$), while utilizing $50$ estimators and a fixed random seed for reproducibility. 

The Decision Tree classifier was optimized with a maximum depth of $15$ to mitigate overfitting, incorporated weights inversely proportional to class frequencies in the input data to address potential class imbalances, and used the Gini impurity criterion as a loss function. 

For LR, the limited-memory Broyden–Fletcher–Goldfarb–Shanno optimization algorithm was selected for efficient convergence on smaller sets, with a maximum of $300$ iterations to ensure optimization stability. Similarly, weights are inversely proportional to class frequencies in the input data for imbalance compensation, and parallel computation is enabled with jobs$=2$. 

The MLP classifier featured a single hidden layer with $50$ neurons, ReLU activation by default, and the Adam optimizer to balance convergence speed and stability, with $300$ maximum iterations to accommodate complex training dynamics. 

All models were integrated into a preprocessing pipeline comprising a univariate imputer for completing missing value handling and a robust scaler for outlier-resistant feature normalization. Across models, a fixed random state ensured deterministic training outcomes, critical for comparative analysis of synthetic data quality. Hyperparameter selections emphasized interpretability (e.g., limited tree depth), computational efficiency (e.g., parallelized LR), and robustness to dataset artifacts inherent in synthetic data evaluation scenarios.
\section*{Conflict of Interest and Data Availability}
The authors declare the following financial interests/personal relationships which may be considered as potential competing interests: Not Applicable. Furthermore, both datasets used are available at https://github.com/Inars/Obsolescence-Forecasting-and-Data-Augmentation-Using-Deep-Generative-Models.
\bibliographystyle{elsarticle-num-names} 
\bibliography{references}

\begin{thebibliography}{58}
\expandafter\ifx\csname natexlab\endcsname\relax\def\natexlab#1{#1}\fi
\providecommand{\url}[1]{\texttt{#1}}
\providecommand{\href}[2]{#2}
\providecommand{\path}[1]{#1}
\providecommand{\DOIprefix}{doi:}
\providecommand{\ArXivprefix}{arXiv:}
\providecommand{\URLprefix}{URL: }
\providecommand{\Pubmedprefix}{pmid:}
\providecommand{\doi}[1]{\href{http://dx.doi.org/#1}{\path{#1}}}
\providecommand{\Pubmed}[1]{\href{pmid:#1}{\path{#1}}}
\providecommand{\bibinfo}[2]{#2}
\ifx\xfnm\relax \def\xfnm[#1]{\unskip,\space#1}\fi
%Type = Book
\bibitem[{DoD(2022)}]{SD22}
\bibinfo{author}{D.~S. P.~O. DoD}, \bibinfo{title}{Diminishing Manufacturing Sources and Material Shortages A Guidebook of Best Practices for Implementing a Robust DMSMS Management Program}, \bibinfo{publisher}{United States Department of Defense}, \bibinfo{year}{2022}.
%Type = Article
\bibitem[{Trabelsi et~al.(2021)Trabelsi, Zolghadri, Zeddini, Barkallah, and Haddar}]{trabelsi2021prediction}
\bibinfo{author}{I.~Trabelsi}, \bibinfo{author}{M.~Zolghadri}, \bibinfo{author}{B.~Zeddini}, \bibinfo{author}{M.~Barkallah}, \bibinfo{author}{M.~Haddar},
\newblock \bibinfo{title}{Prediction of obsolescence degree as a function of time: A mathematical formulation},
\newblock \bibinfo{journal}{Computers in Industry} \bibinfo{volume}{129} (\bibinfo{year}{2021}) \bibinfo{pages}{103470}.
%Type = Article
\bibitem[{Rojo et~al.(2012)Rojo, Baguley, Shaikh, Roy, and Kelly}]{rojo2012}
\bibinfo{author}{F.~J.~R. Rojo}, \bibinfo{author}{P.~Baguley}, \bibinfo{author}{N.~Shaikh}, \bibinfo{author}{R.~Roy}, \bibinfo{author}{S.~Kelly},
\newblock \bibinfo{title}{Tomcat: An obsolescence management capability assessment framework},
\newblock \bibinfo{journal}{Journal of Physics: Conference Series} \bibinfo{volume}{364} (\bibinfo{year}{2012}) \bibinfo{pages}{012098}. \URLprefix \url{https://dx.doi.org/10.1088/1742-6596/364/1/012098}. \DOIprefix\doi{10.1088/1742-6596/364/1/012098}.
%Type = Techreport
\bibitem[{IEC 62402:2019(2019)}]{IEC62402}
IEC 62402:2019, \bibinfo{title}{{Obsolescence management}}, \bibinfo{type}{Standard}, International Electrotechnical Commission, \bibinfo{address}{Geneva, CH}, \bibinfo{year}{2019}.
%Type = Article
\bibitem[{Zolghadri et~al.(2023)Zolghadri, Besbes, Bourgeois, and Saad}]{zolghadri2023micro}
\bibinfo{author}{M.~Zolghadri}, \bibinfo{author}{M.~Besbes}, \bibinfo{author}{V.~Bourgeois}, \bibinfo{author}{E.~Saad},
\newblock \bibinfo{title}{Micro-electronic chips shortages and obsolescence: an empirical study},
\newblock \bibinfo{journal}{Procedia CIRP} \bibinfo{volume}{120} (\bibinfo{year}{2023}) \bibinfo{pages}{1570--1575}.
%Type = Article
\bibitem[{Mellal(2020)}]{mellal2020obsolescence}
\bibinfo{author}{M.~A. Mellal},
\newblock \bibinfo{title}{Obsolescence--a review of the literature},
\newblock \bibinfo{journal}{Technology in Society} \bibinfo{volume}{63} (\bibinfo{year}{2020}) \bibinfo{pages}{101347}.
%Type = Article
\bibitem[{Zolghadri et~al.(2021)Zolghadri, Addouche, Baron, Soltan, and Boissie}]{zolghadri2021obsolescence}
\bibinfo{author}{M.~Zolghadri}, \bibinfo{author}{S.-A. Addouche}, \bibinfo{author}{C.~Baron}, \bibinfo{author}{A.~Soltan}, \bibinfo{author}{K.~Boissie},
\newblock \bibinfo{title}{Obsolescence, rarefaction and their propagation},
\newblock \bibinfo{journal}{Research in Engineering Design} \bibinfo{volume}{32} (\bibinfo{year}{2021}) \bibinfo{pages}{451--468}.
%Type = Article
\bibitem[{{\'Z}r{\'o}bek(2011)}]{zrobek2011remarks}
\bibinfo{author}{R.~{\'Z}r{\'o}bek},
\newblock \bibinfo{title}{Remarks about methods of recognizing types of depreciation and obsolescence},
\newblock \bibinfo{journal}{Studia i Materia{\l}y Towarzystwa Naukowego Nieruchomo{\'s}ci}  (\bibinfo{year}{2011}) \bibinfo{pages}{65--72}.
%Type = Article
\bibitem[{Butt et~al.(2015)Butt, Camilleri, Paul, and Jones}]{Butt2015ObsolescenceTA}
\bibinfo{author}{T.~E. Butt}, \bibinfo{author}{M.~Camilleri}, \bibinfo{author}{P.~Paul}, \bibinfo{author}{K.~A. Jones},
\newblock \bibinfo{title}{Obsolescence types and the built environment - definitions and implications},
\newblock \bibinfo{journal}{International Journal of Environment and Sustainable Development} \bibinfo{volume}{14} (\bibinfo{year}{2015}) \bibinfo{pages}{20}.
%Type = Article
\bibitem[{Rust et~al.(2022)Rust, Elshennawy, and Rabelo}]{rust2022literature}
\bibinfo{author}{R.~M. Rust}, \bibinfo{author}{A.~Elshennawy}, \bibinfo{author}{L.~Rabelo},
\newblock \bibinfo{title}{A literature review on mitigation strategies for electrical component obsolescence in military-based systems},
\newblock \bibinfo{journal}{South African Journal of Industrial Engineering} \bibinfo{volume}{33} (\bibinfo{year}{2022}) \bibinfo{pages}{25--38}.
%Type = Article
\bibitem[{Jenab et~al.(2014)Jenab, Noori, and Weinsier}]{Jenab2014ObsolescenceMI}
\bibinfo{author}{K.~Jenab}, \bibinfo{author}{K.~Noori}, \bibinfo{author}{P.~D. Weinsier},
\newblock \bibinfo{title}{Obsolescence management in rail signalling systems: concept and markovian modelling},
\newblock \bibinfo{journal}{International Journal of Productivity and Quality Management} \bibinfo{volume}{14} (\bibinfo{year}{2014}) \bibinfo{pages}{21--35}.
%Type = Article
\bibitem[{Francesco et~al.(2019)Francesco, Francesco, and Leccese}]{DeFrancesco2019UseOT}
\bibinfo{author}{E.~D. Francesco}, \bibinfo{author}{R.~D. Francesco}, \bibinfo{author}{F.~Leccese},
\newblock \bibinfo{title}{Use of the asd s3000l for the optimization of projects in order to reduce the risk of obsolescence of complex systems},
\newblock \bibinfo{journal}{2019 IEEE 5th International Workshop on Metrology for AeroSpace (MetroAeroSpace)}  (\bibinfo{year}{2019}) \bibinfo{pages}{233--237}.
%Type = Article
\bibitem[{Riascos et~al.(2019)Riascos, Wang-Michelitsch, and Michelitsch}]{riascos2019aging}
\bibinfo{author}{A.~P. Riascos}, \bibinfo{author}{J.~Wang-Michelitsch}, \bibinfo{author}{T.~Michelitsch},
\newblock \bibinfo{title}{Aging in transport processes on networks with stochastic cumulative damage},
\newblock \bibinfo{journal}{Physical Review E} \bibinfo{volume}{100} (\bibinfo{year}{2019}) \bibinfo{pages}{022312}.
%Type = Article
\bibitem[{Tchuente et~al.(2024)Tchuente, Lonlac, and Kamsu-Foguem}]{tchuente2024methodological}
\bibinfo{author}{D.~Tchuente}, \bibinfo{author}{J.~Lonlac}, \bibinfo{author}{B.~Kamsu-Foguem},
\newblock \bibinfo{title}{A methodological and theoretical framework for implementing explainable artificial intelligence (xai) in business applications},
\newblock \bibinfo{journal}{Computers in Industry} \bibinfo{volume}{155} (\bibinfo{year}{2024}) \bibinfo{pages}{104044}.
%Type = Article
\bibitem[{Culot et~al.(2024)Culot, Podrecca, and Nassimbeni}]{culot2024artificial}
\bibinfo{author}{G.~Culot}, \bibinfo{author}{M.~Podrecca}, \bibinfo{author}{G.~Nassimbeni},
\newblock \bibinfo{title}{Artificial intelligence in supply chain management: A systematic literature review of empirical studies and research directions},
\newblock \bibinfo{journal}{Computers in Industry} \bibinfo{volume}{162} (\bibinfo{year}{2024}) \bibinfo{pages}{104132}.
%Type = Article
\bibitem[{Trabelsi et~al.(2021)Trabelsi, Zeddini, Zolghadri, Barkallah, and Haddar}]{trabelsi2021obsolescence}
\bibinfo{author}{I.~Trabelsi}, \bibinfo{author}{B.~Zeddini}, \bibinfo{author}{M.~Zolghadri}, \bibinfo{author}{M.~Barkallah}, \bibinfo{author}{M.~Haddar},
\newblock \bibinfo{title}{Obsolescence prediction based on joint feature selection and machine learning techniques.},
\newblock \bibinfo{journal}{ICAART (2)}  (\bibinfo{year}{2021}) \bibinfo{pages}{787--794}.
%Type = Article
\bibitem[{Jennings et~al.(2016)Jennings, Wu, and Terpenny}]{jennings2016forecasting}
\bibinfo{author}{C.~Jennings}, \bibinfo{author}{D.~Wu}, \bibinfo{author}{J.~Terpenny},
\newblock \bibinfo{title}{Forecasting obsolescence risk and product life cycle with machine learning},
\newblock \bibinfo{journal}{IEEE Transactions on Components, Packaging and Manufacturing Technology} \bibinfo{volume}{6} (\bibinfo{year}{2016}) \bibinfo{pages}{1428--1439}.
%Type = Article
\bibitem[{Moon et~al.(2022)Moon, Lee, and Kim}]{moon2022adaptive}
\bibinfo{author}{K.-S. Moon}, \bibinfo{author}{H.~W. Lee}, \bibinfo{author}{H.~Kim},
\newblock \bibinfo{title}{Adaptive data selection-based machine learning algorithm for prediction of component obsolescence},
\newblock \bibinfo{journal}{Sensors} \bibinfo{volume}{22} (\bibinfo{year}{2022}) \bibinfo{pages}{7982}.
%Type = Book
\bibitem[{Zhou(2021)}]{zhou2021machine}
\bibinfo{author}{Z.-H. Zhou}, \bibinfo{title}{Machine learning}, \bibinfo{publisher}{Springer nature}, \bibinfo{year}{2021}.
%Type = Inproceedings
\bibitem[{Hubauer et~al.(2013)Hubauer, Lamparter, Roshchin, Solomakhina, and Watson}]{hubauer2013analysis}
\bibinfo{author}{T.~Hubauer}, \bibinfo{author}{S.~Lamparter}, \bibinfo{author}{M.~Roshchin}, \bibinfo{author}{N.~Solomakhina}, \bibinfo{author}{S.~Watson},
\newblock \bibinfo{title}{Analysis of data quality issues in real-world industrial data},
\newblock in: \bibinfo{booktitle}{Annual Conference of the PHM Society}, volume~\bibinfo{volume}{5}, \bibinfo{year}{2013}.
%Type = Inproceedings
\bibitem[{Jess et~al.(2015)Jess, Woodall, and McFarlane}]{jess2015overcoming}
\bibinfo{author}{T.~Jess}, \bibinfo{author}{P.~Woodall}, \bibinfo{author}{D.~McFarlane},
\newblock \bibinfo{title}{Overcoming limited dataset availability when working with industrial organisations},
\newblock in: \bibinfo{booktitle}{2015 IEEE 13th International Conference on Industrial Informatics (INDIN)}, \bibinfo{organization}{IEEE}, \bibinfo{year}{2015}, pp. \bibinfo{pages}{826--831}.
%Type = Inproceedings
\bibitem[{Libes et~al.(2015)Libes, Shin, and Woo}]{libes2015considerations}
\bibinfo{author}{D.~Libes}, \bibinfo{author}{S.~Shin}, \bibinfo{author}{J.~Woo},
\newblock \bibinfo{title}{Considerations and recommendations for data availability for data analytics for manufacturing},
\newblock in: \bibinfo{booktitle}{2015 IEEE International Conference on Big Data (Big Data)}, \bibinfo{organization}{IEEE}, \bibinfo{year}{2015}, pp. \bibinfo{pages}{68--75}.
%Type = Inproceedings
\bibitem[{Grichi et~al.(2017)Grichi, Beauregard, and Dao}]{grichi2017random}
\bibinfo{author}{Y.~Grichi}, \bibinfo{author}{Y.~Beauregard}, \bibinfo{author}{T.~Dao},
\newblock \bibinfo{title}{A random forest method for obsolescence forecasting},
\newblock in: \bibinfo{booktitle}{2017 IEEE International Conference on Industrial Engineering and Engineering Management (IEEM)}, \bibinfo{organization}{IEEE}, \bibinfo{year}{2017}, pp. \bibinfo{pages}{1602--1606}.
%Type = Article
\bibitem[{Liu and Zhao(2022)}]{liu2022obsolescence}
\bibinfo{author}{Y.~Liu}, \bibinfo{author}{M.~Zhao},
\newblock \bibinfo{title}{An obsolescence forecasting method based on improved radial basis function neural network},
\newblock \bibinfo{journal}{Ain Shams Engineering Journal} \bibinfo{volume}{13} (\bibinfo{year}{2022}) \bibinfo{pages}{101775}.
%Type = Article
\bibitem[{Moon et~al.(2022)Moon, Lee, Kim, Kim, Kang, and Paik}]{moon2022forecasting}
\bibinfo{author}{K.-S. Moon}, \bibinfo{author}{H.~W. Lee}, \bibinfo{author}{H.~J. Kim}, \bibinfo{author}{H.~Kim}, \bibinfo{author}{J.~Kang}, \bibinfo{author}{W.~C. Paik},
\newblock \bibinfo{title}{Forecasting obsolescence of components by using a clustering-based hybrid machine-learning algorithm},
\newblock \bibinfo{journal}{Sensors} \bibinfo{volume}{22} (\bibinfo{year}{2022}) \bibinfo{pages}{3244}.
%Type = Article
\bibitem[{Cholaquidis et~al.(2020)Cholaquidis, Fraiman, and Sued}]{cholaquidis2020semi}
\bibinfo{author}{A.~Cholaquidis}, \bibinfo{author}{R.~Fraiman}, \bibinfo{author}{M.~Sued},
\newblock \bibinfo{title}{On semi-supervised learning},
\newblock \bibinfo{journal}{TEST} \bibinfo{volume}{29} (\bibinfo{year}{2020}) \bibinfo{pages}{914--937}.
%Type = Article
\bibitem[{Breiman(2001)}]{breiman2001random}
\bibinfo{author}{L.~Breiman},
\newblock \bibinfo{title}{Random forests},
\newblock \bibinfo{journal}{Machine learning} \bibinfo{volume}{45} (\bibinfo{year}{2001}) \bibinfo{pages}{5--32}.
%Type = Article
\bibitem[{Grichi et~al.(2018{\natexlab{a}})Grichi, Beauregard, and Dao}]{grichi2018optimization}
\bibinfo{author}{Y.~Grichi}, \bibinfo{author}{Y.~Beauregard}, \bibinfo{author}{T.-M. Dao},
\newblock \bibinfo{title}{Optimization of obsolescence forecasting using new hybrid approach based on the rf method and the meta-heuristic genetic algorithm},
\newblock \bibinfo{journal}{American Journal of Management} \bibinfo{volume}{18} (\bibinfo{year}{2018}{\natexlab{a}}).
%Type = Inproceedings
\bibitem[{Grichi et~al.(2018{\natexlab{b}})Grichi, Dao, and Beauregard}]{grichi2018new}
\bibinfo{author}{Y.~Grichi}, \bibinfo{author}{T.-M. Dao}, \bibinfo{author}{Y.~Beauregard},
\newblock \bibinfo{title}{A new approach for optimal obsolescence forecasting based on the random forest (rf) technique and meta-heuristic particle swarm optimization (pso)},
\newblock in: \bibinfo{booktitle}{Proceedings of the International Conference on Industrial Engineering and Operations Management, Paris, France}, \bibinfo{year}{2018}{\natexlab{b}}, pp. \bibinfo{pages}{26--27}.
%Type = Misc
\bibitem[{Sierra-Fontalvo et~al.(2023)Sierra-Fontalvo, Gonzalez-Quiroga, and Mesa}]{sierra2023deep}
\bibinfo{author}{L.~Sierra-Fontalvo}, \bibinfo{author}{A.~Gonzalez-Quiroga}, \bibinfo{author}{J.~Mesa}, \bibinfo{title}{A deep dive into addressing obsolescence in product design: A review. heliyon, 9 (11), e21856}, \bibinfo{year}{2023}.
%Type = Misc
\bibitem[{Goodfellow(2016)}]{goodfellow2016deep}
\bibinfo{author}{I.~Goodfellow}, \bibinfo{title}{Deep learning}, \bibinfo{year}{2016}.
%Type = Article
\bibitem[{Nassef et~al.(2023)Nassef, Abdelkareem, Maghrabie, and Baroutaji}]{nassef2023review}
\bibinfo{author}{A.~M. Nassef}, \bibinfo{author}{M.~A. Abdelkareem}, \bibinfo{author}{H.~M. Maghrabie}, \bibinfo{author}{A.~Baroutaji},
\newblock \bibinfo{title}{Review of metaheuristic optimization algorithms for power systems problems},
\newblock \bibinfo{journal}{Sustainability} \bibinfo{volume}{15} (\bibinfo{year}{2023}) \bibinfo{pages}{9434}.
%Type = Article
\bibitem[{Dinh et~al.(2016)Dinh, Sohl-Dickstein, and Bengio}]{dinh2016density}
\bibinfo{author}{L.~Dinh}, \bibinfo{author}{J.~Sohl-Dickstein}, \bibinfo{author}{S.~Bengio},
\newblock \bibinfo{title}{Density estimation using real nvp},
\newblock \bibinfo{journal}{arXiv preprint arXiv:1605.08803}  (\bibinfo{year}{2016}).
%Type = Phdthesis
\bibitem[{Xu et~al.(2020)}]{xu2020synthesizing}
\bibinfo{author}{L.~Xu}, et~al., \bibinfo{title}{Synthesizing tabular data using conditional GAN}, Ph.D. thesis, Massachusetts Institute of Technology, \bibinfo{year}{2020}.
%Type = Article
\bibitem[{Duff et~al.(2024)Duff, Campbell, and Ehrhardt}]{duff2024regularising}
\bibinfo{author}{M.~Duff}, \bibinfo{author}{N.~D. Campbell}, \bibinfo{author}{M.~J. Ehrhardt},
\newblock \bibinfo{title}{Regularising inverse problems with generative machine learning models},
\newblock \bibinfo{journal}{Journal of Mathematical Imaging and Vision} \bibinfo{volume}{66} (\bibinfo{year}{2024}) \bibinfo{pages}{37--56}.
%Type = Article
\bibitem[{Srinivas and Babu(2015)}]{srinivas2015deep}
\bibinfo{author}{S.~Srinivas}, \bibinfo{author}{R.~V. Babu},
\newblock \bibinfo{title}{Deep learning in neural networks: An overview},
\newblock \bibinfo{journal}{Computer Science}  (\bibinfo{year}{2015}).
%Type = Article
\bibitem[{Srivastava et~al.(2014)Srivastava, Hinton, Krizhevsky, Sutskever, and Salakhutdinov}]{srivastava2014dropout}
\bibinfo{author}{N.~Srivastava}, \bibinfo{author}{G.~Hinton}, \bibinfo{author}{A.~Krizhevsky}, \bibinfo{author}{I.~Sutskever}, \bibinfo{author}{R.~Salakhutdinov},
\newblock \bibinfo{title}{Dropout: a simple way to prevent neural networks from overfitting},
\newblock \bibinfo{journal}{The journal of machine learning research} \bibinfo{volume}{15} (\bibinfo{year}{2014}) \bibinfo{pages}{1929--1958}.
%Type = Inproceedings
\bibitem[{Janakiramaiah et~al.(2020)Janakiramaiah, Kalyani, Narayana, and Bala Murali~Krishna}]{janakiramaiah2020reducing}
\bibinfo{author}{B.~Janakiramaiah}, \bibinfo{author}{G.~Kalyani}, \bibinfo{author}{S.~Narayana}, \bibinfo{author}{T.~Bala Murali~Krishna},
\newblock \bibinfo{title}{Reducing dimensionality of data using autoencoders},
\newblock in: \bibinfo{booktitle}{Smart Intelligent Computing and Applications: Proceedings of the Third International Conference on Smart Computing and Informatics, Volume 2}, \bibinfo{organization}{Springer}, \bibinfo{year}{2020}, pp. \bibinfo{pages}{51--58}.
%Type = Article
\bibitem[{Wang et~al.(2016)Wang, Yao, and Zhao}]{wang2016auto}
\bibinfo{author}{Y.~Wang}, \bibinfo{author}{H.~Yao}, \bibinfo{author}{S.~Zhao},
\newblock \bibinfo{title}{Auto-encoder based dimensionality reduction},
\newblock \bibinfo{journal}{Neurocomputing} \bibinfo{volume}{184} (\bibinfo{year}{2016}) \bibinfo{pages}{232--242}.
%Type = Inproceedings
\bibitem[{Fournier and Aloise(2019)}]{fournier2019empirical}
\bibinfo{author}{Q.~Fournier}, \bibinfo{author}{D.~Aloise},
\newblock \bibinfo{title}{Empirical comparison between autoencoders and traditional dimensionality reduction methods},
\newblock in: \bibinfo{booktitle}{2019 IEEE Second International Conference on Artificial Intelligence and Knowledge Engineering (AIKE)}, \bibinfo{organization}{IEEE}, \bibinfo{year}{2019}, pp. \bibinfo{pages}{211--214}.
%Type = Article
\bibitem[{Amini et~al.(2024)Amini, Feofanov, Pauletto, Hadjadj, Devijver, and Maximov}]{amini2022self}
\bibinfo{author}{M.-R. Amini}, \bibinfo{author}{V.~Feofanov}, \bibinfo{author}{L.~Pauletto}, \bibinfo{author}{L.~Hadjadj}, \bibinfo{author}{E.~Devijver}, \bibinfo{author}{Y.~Maximov},
\newblock \bibinfo{title}{Self-training: A survey},
\newblock \bibinfo{journal}{arXiv preprint arXiv:2202.12040}  (\bibinfo{year}{2024}).
%Type = Article
\bibitem[{Chai and Draxler(2014)}]{chai2014root}
\bibinfo{author}{T.~Chai}, \bibinfo{author}{R.~R. Draxler},
\newblock \bibinfo{title}{Root mean square error (rmse) or mean absolute error (mae)?--arguments against avoiding rmse in the literature},
\newblock \bibinfo{journal}{Geoscientific model development} \bibinfo{volume}{7} (\bibinfo{year}{2014}) \bibinfo{pages}{1247--1250}.
%Type = Inproceedings
\bibitem[{Carrara and Ernst(2020)}]{carrara2020estimation}
\bibinfo{author}{N.~Carrara}, \bibinfo{author}{J.~Ernst},
\newblock \bibinfo{title}{On the estimation of mutual information},
\newblock in: \bibinfo{booktitle}{Proceedings}, volume~\bibinfo{volume}{33}, \bibinfo{organization}{MDPI}, \bibinfo{year}{2020}, p.~\bibinfo{pages}{31}.
%Type = Article
\bibitem[{Baez et~al.(2011)Baez, Fritz, and Leinster}]{baez2011characterization}
\bibinfo{author}{J.~C. Baez}, \bibinfo{author}{T.~Fritz}, \bibinfo{author}{T.~Leinster},
\newblock \bibinfo{title}{A characterization of entropy in terms of information loss},
\newblock \bibinfo{journal}{Entropy} \bibinfo{volume}{13} (\bibinfo{year}{2011}) \bibinfo{pages}{1945--1957}.
%Type = Article
\bibitem[{Chakraborty(2022)}]{chakraborty2022topsis}
\bibinfo{author}{S.~Chakraborty},
\newblock \bibinfo{title}{Topsis and modified topsis: A comparative analysis},
\newblock \bibinfo{journal}{Decision Analytics Journal} \bibinfo{volume}{2} (\bibinfo{year}{2022}) \bibinfo{pages}{100021}.
%Type = Article
\bibitem[{Hodges~Jr(1958)}]{hodges1958significance}
\bibinfo{author}{J.~Hodges~Jr},
\newblock \bibinfo{title}{The significance probability of the smirnov two-sample test},
\newblock \bibinfo{journal}{Arkiv f{\"o}r matematik} \bibinfo{volume}{3} (\bibinfo{year}{1958}) \bibinfo{pages}{469--486}.
%Type = Article
\bibitem[{Berman(2016)}]{berman2016chapter}
\bibinfo{author}{J.~J. Berman},
\newblock \bibinfo{title}{Chapter 4—understanding your data},
\newblock \bibinfo{journal}{Data simplification}  (\bibinfo{year}{2016}) \bibinfo{pages}{135--187}.
%Type = Article
\bibitem[{Taboga(2017)}]{taboga2017lectures}
\bibinfo{author}{M.~Taboga},
\newblock \bibinfo{title}{Lectures on probability theory and mathematical statistics},
\newblock \bibinfo{journal}{(No Title)}  (\bibinfo{year}{2017}).
%Type = Article
\bibitem[{Xu et~al.(2023)Xu, Sun, and Cheng}]{xu2023utility}
\bibinfo{author}{S.~Xu}, \bibinfo{author}{W.~W. Sun}, \bibinfo{author}{G.~Cheng},
\newblock \bibinfo{title}{Utility theory of synthetic data generation},
\newblock \bibinfo{journal}{arXiv preprint arXiv:2305.10015}  (\bibinfo{year}{2023}).
%Type = Inbook
\bibitem[{Shalev-Shwartz and Ben-David(2014)}]{Shalev-Shwartz_Ben-David_2014}
\bibinfo{author}{S.~Shalev-Shwartz}, \bibinfo{author}{S.~Ben-David}, \bibinfo{title}{Decision Trees}, \bibinfo{publisher}{Cambridge University Press}, \bibinfo{year}{2014}, p. \bibinfo{pages}{212–218}.
%Type = Article
\bibitem[{Hastie et~al.(2009)Hastie, Rosset, Zhu, and Zou}]{hastie2009multi}
\bibinfo{author}{T.~Hastie}, \bibinfo{author}{S.~Rosset}, \bibinfo{author}{J.~Zhu}, \bibinfo{author}{H.~Zou},
\newblock \bibinfo{title}{Multi-class adaboost},
\newblock \bibinfo{journal}{Statistics and its Interface} \bibinfo{volume}{2} (\bibinfo{year}{2009}) \bibinfo{pages}{349--360}.
%Type = Misc
\bibitem[{Saad(2024)}]{saad_2024_15017365}
\bibinfo{author}{E.~Saad}, \bibinfo{title}{Zenner diod obsolescence dataset}, \bibinfo{year}{2024}. \URLprefix \url{https://doi.org/10.5281/zenodo.15017365}. \DOIprefix\doi{10.5281/zenodo.15017365}.
%Type = Article
\bibitem[{Smirnov(1948)}]{smirnov1948table}
\bibinfo{author}{N.~Smirnov},
\newblock \bibinfo{title}{Table for estimating the goodness of fit of empirical distributions},
\newblock \bibinfo{journal}{The annals of mathematical statistics} \bibinfo{volume}{19} (\bibinfo{year}{1948}) \bibinfo{pages}{279--281}.
%Type = Book
\bibitem[{Hastie et~al.(2009)Hastie, Tibshirani, Friedman, and Friedman}]{hastie2009elements}
\bibinfo{author}{T.~Hastie}, \bibinfo{author}{R.~Tibshirani}, \bibinfo{author}{J.~H. Friedman}, \bibinfo{author}{J.~H. Friedman}, \bibinfo{title}{The elements of statistical learning: data mining, inference, and prediction}, volume~\bibinfo{volume}{2}, \bibinfo{publisher}{Springer}, \bibinfo{year}{2009}.
%Type = Article
\bibitem[{Junge and Dettori(2018)}]{junge2018roc}
\bibinfo{author}{M.~R. Junge}, \bibinfo{author}{J.~R. Dettori},
\newblock \bibinfo{title}{Roc solid: Receiver operator characteristic (roc) curves as a foundation for better diagnostic tests},
\newblock \bibinfo{journal}{Global Spine Journal} \bibinfo{volume}{8} (\bibinfo{year}{2018}) \bibinfo{pages}{424--429}.
%Type = Inproceedings
\bibitem[{Calders and Jaroszewicz(2007)}]{calders2007efficient}
\bibinfo{author}{T.~Calders}, \bibinfo{author}{S.~Jaroszewicz},
\newblock \bibinfo{title}{Efficient auc optimization for classification},
\newblock in: \bibinfo{booktitle}{European conference on principles of data mining and knowledge discovery}, \bibinfo{organization}{Springer}, \bibinfo{year}{2007}, pp. \bibinfo{pages}{42--53}.
%Type = Article
\bibitem[{Papamakarios et~al.(2021)Papamakarios, Nalisnick, Rezende, Mohamed, and Lakshminarayanan}]{papamakarios2021normalizing}
\bibinfo{author}{G.~Papamakarios}, \bibinfo{author}{E.~Nalisnick}, \bibinfo{author}{D.~J. Rezende}, \bibinfo{author}{S.~Mohamed}, \bibinfo{author}{B.~Lakshminarayanan},
\newblock \bibinfo{title}{Normalizing flows for probabilistic modeling and inference},
\newblock \bibinfo{journal}{Journal of Machine Learning Research} \bibinfo{volume}{22} (\bibinfo{year}{2021}) \bibinfo{pages}{1--64}.
%Type = Article
\bibitem[{Kobyzev et~al.(2020)Kobyzev, Prince, and Brubaker}]{kobyzev2020normalizing}
\bibinfo{author}{I.~Kobyzev}, \bibinfo{author}{S.~J. Prince}, \bibinfo{author}{M.~A. Brubaker},
\newblock \bibinfo{title}{Normalizing flows: An introduction and review of current methods},
\newblock \bibinfo{journal}{IEEE transactions on pattern analysis and machine intelligence} \bibinfo{volume}{43} (\bibinfo{year}{2020}) \bibinfo{pages}{3964--3979}.

\end{thebibliography}

\end{document}